\documentclass[accepted]{uai2026} 
                        
\usepackage[american]{babel}

\usepackage{natbib} 
\usepackage{mathtools} 
\usepackage{booktabs} 
\usepackage{tikz} 
\usepackage{enumitem}

\usepackage{algorithm}
\usepackage{algorithmic}

\usepackage[capitalize,noabbrev]{cleveref}
\usepackage[utf8]{inputenc}
\usepackage{graphicx}      
\usepackage{booktabs}      

\usepackage{amssymb}

\crefname{equation}{Eq.}{Eqs.}

\newcommand \indep{\mathop{\perp\!\!\!\!\perp}}

\DeclareMathOperator{\E}{\mathbb{E}}

\DeclareMathOperator{\R}{\mathbb{R}}
\DeclareMathOperator{\pr}{\mathrm{P}}

\newcommand{\biX}{\textbf{\textit{X}}}
\newcommand{\bix}{\textbf{\textit{x}}}

\newcommand{\bim}{\textbf{\textit{m}}}
\newcommand{\cg}{\mathcal{G}}

\newcommand{\biA}{\mathbf{A}}
\newcommand{\bitA}{\textbf{\textit{A}}}
\newcommand{\biU}{\mathbf{U}}
\newcommand{\biPi}{\mathbf{\Pi}}

\newcommand{\biphi}{\boldsymbol{\phi}}

\newcommand{\biF}{\textbf{\textit{F}}}
\newcommand{\bic}{\textbf{\textit{c}}}
\newcommand{\biw}{\textbf{\textit{w}}}

\usepackage{amsthm}
\usepackage{thm-restate}
\usepackage{thmtools} 
\theoremstyle{plain}
\declaretheoremstyle[
  headpunct={:},
  headfont=\bfseries,          
  bodyfont=\itshape,           
  notefont=\normalfont,  
]{amsthmstyle}

\declaretheorem[
  style=amsthmstyle,
  name=Theorem,
  numberwithin=section
]{theorem}

\theoremstyle{definition}

\theoremstyle{remark}
\newtheorem{remark}[theorem]{Remark}

\title{MetaCaDI: A Meta-Learning Framework for Causal Discovery\\from Multiple Environments with Unknown Interventions}

\author[1]{\href{mailto:ong.hans_jarett.ol5@naist.ac.jp?Subject=Your UAI 2026 paper}{Hans Jarett Ong}{}}
\author[2]{Yoichi Chikahara}
\author[2]{Tomoharu Iwata}
\affil[1]{%
    Nara Institute of Science and Technology\\
    Nara, Japan
}
\affil[2]{%
    NTT Communication Science Laboratories\\
    Kyoto, Japan\\
}
  
\begin{document}
\maketitle

\begin{abstract}
Uncovering the causal mechanisms of complex real-world systems remains a significant challenge, as these systems often entail high data collection costs and involve unknown interventions. We introduce MetaCaDI, the first framework to cast the identification of unknown interventions as a meta-learning problem, explicitly leveraging a jointly learned causal graph. MetaCaDI is a Bayesian framework that learns a shared causal structure across multiple environments and is optimized to rapidly adapt to new, few-shot intervention target identification tasks. A key innovation is our model's analytical adaptation, which uses a closed-form solution to bypass expensive and potentially unstable gradient-based bilevel optimization. Extensive experiments on synthetic and complex gene expression data demonstrate that MetaCaDI significantly outperforms state-of-the-art methods. It excels at identifying intervention targets from as few as 3 samples—where existing methods collapse to random chance—while robustly recovering the shared causal graph, proving its effectiveness in data-scarce scenarios.
\end{abstract}

\section{Introduction}
\label{sec:intro}

This paper addresses the problem of causal discovery from multiple environments, which has emerged as a promising avenue for unraveling complex real-world systems \citep{jci_mooij_jmlr2020,Li2023_multienv_causal_discovery,jalaldoust25a_multidomain_causal_discovery}. In this problem, we consider the setting where we have access to multiple datasets generated by a shared causal graph structure, but each dataset is produced from a different \textit{environment}, where distinct variables are intervened upon. The goal of this problem is to jointly infer the shared causal graph structure and distinct, unknown intervention targets.

Although this problem is pervasive in real-world applications, achieving high performance remains challenging, as the data sample size is often limited in practice. To illustrate this, we present two motivating examples in industrial and medical applications:

\paragraph{Example 1 (Large computing system):} 
Consider a large-scale cloud computing system whose components (e.g., physical servers and software applications) occasionally experience localized faults such as hardware malfunctions or software bugs \citep{varici22a_interv_example}. We can represent such a system as a causal graph whose nodes represent the components and whose edges express the dependencies among them. Due to the complex interdependencies, a system engineer wants to identify the graph structure and the causes of the faults (i.e., the intervention targets) \citep{Mariani2018LocalizingFI_interv_example}. The engineer has access to multiple datasets collected under different system conditions. However, such datasets are often scarce, as one cannot intentionally cause a system fault. Nevertheless, the engineer needs to diagnose failures \textbf{in real-time}, using only a few observations of the failure.

\paragraph{Example 2 (Pharmacological neuroscience):}
In neuroscience, it is common to model the human brain as a causal graph where nodes represent brain regions and edges express the effective connectivity among them \citep{friston2003dynamic, valdes2011effective}. By collecting functional magnetic resonance imaging (fMRI) datasets from patient cohorts administered different pharmacological treatments, neurologists seek to infer this graph structure shared across datasets, as well as the specific brain regions modulated by each drug, which can be regarded as an unknown intervention target \citep{rowe2010dynamic, oliva2022simultaneous}. However, patient datasets for highly specific drug combinations or novel treatments are inherently scarce, making large-scale data collection costly. Therefore, identifying the specific brain regions affected by a drug from only a handful of patient records would greatly aid in understanding drug mechanisms and advancing targeted clinical care.


Motivated by these challenging yet important data-scarce scenarios, we propose \textbf{MetaCaDI} (\textit{Meta}-Learning for \textit{Ca}usal \textit{D}iscovery with Unknown \textit{I}nterventions), a meta-learning approach to causal discovery from multiple environments. MetaCaDI is a Bayesian framework for inferring the joint posterior distribution over the shared causal graph and the distinct intervention targets specific to each small dataset. Using the estimated posterior probabilities, we can quantify the inference uncertainty of \textbf{both} the shared graph structure and the distinct intervention targets, which is essential for knowledge discovery under data-scarce scenarios. 

MetaCaDI has two notable strengths. First, unlike multitask learning approaches \citep{bacadi_hagele_2023}, MetaCaDI can address real-time applications like \textbf{Example 1}, as meta-learning does not require the full model retraining to incorporate a newly arrived dataset. Second, MetaCaDI is designed to achieve high meta-learning performance in few-shot settings like \textbf{Example 2} by overcoming the computational inefficiency of standard gradient-based approaches \citep{Bertinetto2018MetalearningWD,maml_finn2017}. To bypass the need for gradient-based updates in few-shot adaptation, we develop an interventional data likelihood model that admits a closed-form, analytical solution, allowing MetaCaDI to rapidly adapt to new datasets with only a handful of samples. 

\textbf{Our contributions} are threefold:
\begin{itemize}[noitemsep, topsep=0pt]
    \item This work is the first to formalize the joint inference of a causal graph and unknown intervention targets as a meta-learning problem (\Cref{sec-problem}). By framing the intervention target identification from each dataset as a distinct meta-learning \textit{task}, our model extracts the \textit{task-shared} knowledge of the causal graph structure
    and then leverages it to improve the performance of inferring \textit{task-specific} intervention targets.

    \item To effectively tune the task-specific parameters for each dataset, we formulate a likelihood model that admits a closed-form, analytical solution (\cref{subsec:modeling}). By eliminating the need for computationally demanding gradient-based optimization for these parameters, this closed-form solver enables computationally efficient and stable adaptation to new datasets.
    
    \item Through extensive experiments on synthetic and gene expression simulation datasets, we demonstrate that MetaCaDI outperforms existing methods in recovering both the causal graph structure and the intervention targets, even in the challenging small-sample setting.
\end{itemize}

\section{Preliminaries}
\label{sec:preliminaries}

\subsection{Causal Graphs and Interventions}
\label{subsec-causal-disc}

A causal graph describes the data-generating process over variables $\mathcal{X} = \{X_1, \dots, X_d\}$, defined by a \textbf{Structural Causal Model (SCM)} \citep{pearl2009_causality_book}.
An SCM defines the \textit{structural equation} for each variable $X_i$ ($i \in \{1, \dots, d\}$), which determines the variable values as the output of a deterministic function $f_i$:
\begin{equation*}
X_i = f_i(\mathrm{PA}(X_i), E_i),
\end{equation*}
where $\mathrm{PA}(X_i) \subseteq \mathcal{X} \backslash X_i$ is a variable subset called the \textit{parents} of $X_i$, 
and $E_i$ is an exogenous variable (e.g., noise). 

A \textbf{causal graph} $\cg$ is a graph that contains a directed edge $X_{j} \rightarrow X_i$ 
if and only if $X_j \in \mathrm{PA}(X_i)$.
We make two standard assumptions on the causal graph $\cg$ and the SCM.
First, we assume that causal graph $\cg$ is a directed acyclic graph (DAG).
Second, we assume \textit{causal sufficiency}, the condition that $E_i \indep E_j$ holds for any $i \neq j$, which requires that there are no unobserved common causes (i.e., confounders) between any pair of variables in $\mathcal{X}$. 
As with the literature in this field \citep{bacadi_hagele_2023,brouillard2020_diff_interv}, to effectively instantiate the likelihood, we make an additional but \textbf{non-mandatory} assumption that each structural equation is expressed as a specific class of functions; in this paper, we consider an \textbf{Additive Noise Model (ANM)} \citep{hoyer2008_anm}:
\begin{equation*}
X_i = f_i(\mathrm{PA}(X_i)) + E_i.
\end{equation*}
Theoretically, data generated under such restricted functional forms as ANMs exhibit distributional asymmetries between $X_i \rightarrow X_j$ and $X_j \rightarrow X_i$, enabling the unique identification of causal graphs \citep{peters2014_resit}. 
In the absence of such functional assumptions, from observational data alone, the causal graph can only be identified up to its \textit{Markov Equivalence Class (MEC)}---a set of causal graphs that encode the same conditional independence relations.

An \textbf{intervention} is a replacement operation for the structural equations. In this paper, we consider both \textbf{hard} and \textbf{soft interventions}: The former replaces the structural equation of variable $X_i$ with constant $X_i=x_i$, whereas the latter modifies the structural equation while keeping its dependence on the values of parents $\text{PA}(X_i)$ \citep{pearl2009_causality_book}.

Such structural equation modification for $X_i$ changes the distribution of its descendants but not that of its ancestors, allowing for the disambiguation of causal directions that are otherwise indistinguishable. Without functional assumptions, an MEC is known to be narrowed down to an Interventional MEC (I-MEC) \citep{Eberhardt2007-IMEC}. 
Combined with the distributional asymmetries under ANMs, interventional data yield a robust source of inference because the asymmetries in the observational data distribution can be easily obscured by sampling variability, particularly when the sample size is small.

\subsection{Related Work}
\label{subsec:existing-methods}

\begin{table}[t]
\centering
\caption{A comparison of methods for causal discovery from multiple environments. Our method, MetaCaDI, is the first to formalize the problem using a meta-learning paradigm, enabling rapid adaptation to unseen environments.}
\label{tab:lit_review}
\resizebox{\columnwidth}{!}{%
\begin{tabular}{@{}lcccc@{}}
\toprule
\textbf{Method} & \textbf{\shortstack{Continuous \\ Optimization}} & \textbf{\shortstack{Uncertainty \\ Quantification}} & \textbf{\shortstack{Few-Shot \\ Adaptation}} \\
\midrule
JCI-PC & $\times$ & $\times$ & $\times$ \\
UT-IGSP & $\times$ & $\times$ & $\times$ \\
DCDI-G & \checkmark & $\times$ & $\times$ \\
BaCaDI& \checkmark & \checkmark & $\times$ \\
\textbf{MetaCaDI (Ours)} & \checkmark & \checkmark & \checkmark \\
\bottomrule
\end{tabular}
}
\end{table}

\Cref{tab:lit_review} presents a summary of methods that have been proposed for causal discovery from multiple environments.

Early methods such as \textbf{JCI-PC} \citep{jci_mooij_jmlr2020} and \textbf{UT-IGSP} \citep{jci_extension_squires2020}, which take \textit{constraint-based} and \textit{score-based} approaches, respectively, suffer from scalability issues, particularly when the number of variables $d$ is large.
\textbf{DCDI-G} \citep{brouillard2020_diff_interv} aims to improve the scalability using continuous optimization.
However, all these methods provide only a single point estimate of the causal graph and cannot quantify the uncertainty inherent in small datasets. This weakness is serious in practice, especially when multiple causal graphs are equally plausible.

A principled way to capture inference uncertainty is through Bayesian inference \citep{eaton_murphy_aistats2007}. 
The state-of-the-art Bayesian approach, \textbf{BaCaDI} \citep{bacadi_hagele_2023}, frames the problem as multitask learning, where it treats \textit{all} datasets as training data to jointly infer a shared causal graph and distinct intervention targets.

However, this multitask learning approach is ill-suited for real-time applications where new data arrives sequentially, as illustrated in \textbf{Example 1} in \cref{sec:intro}. To perform multitask learning in such streaming settings, it requires retraining the entire model from scratch to incorporate newly arrived datasets, which is computationally prohibitive for practical applications. 
This limitation highlights the need for a framework that can rapidly adapt to unseen datasets without full retraining, motivating a \textbf{meta-learning} approach.



\section{Meta-Learning Problem and Inference Targets} \label{sec-problem}


\subsection{Problem Setup}
\label{subsec-problem}

We frame the problem of causal discovery from multiple environments as a meta-learning problem.

\begin{figure}[t]
\centering
\includegraphics[width=\columnwidth]{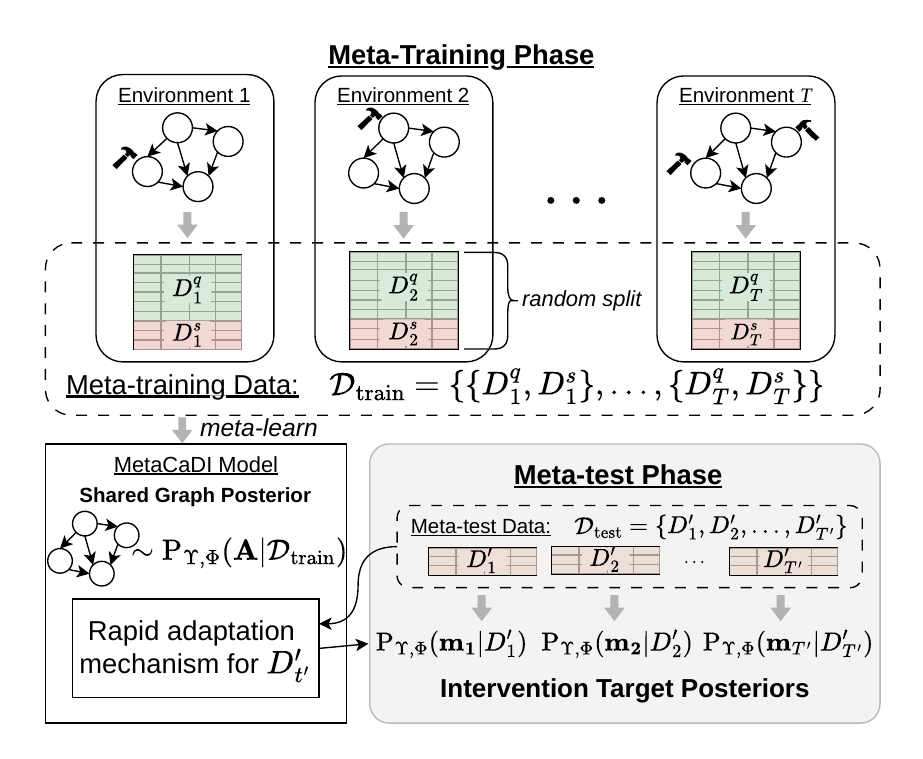}
\caption{Our meta-learning setup. \textbf{Meta-training Phase}: 
Given a collection of datasets $\mathcal{D}_\text{train} = \{D_1, \dots, D_T\}$, MetaCaDI infers shared causal DAG adjacency matrix $\biA$ from $\mathcal{D}_\text{train}$ and intervention target $\bim_t$ for each $D_t$. \textbf{Meta-test Phase}: Using a new, small dataset $D'_{t'}$, MetaCaDI performs fine-tuning to infer its specific intervention targets.
}
\label{fig:prob-setup}
\end{figure}

\Cref{fig:prob-setup} summarizes our meta-learning problem for inferring the shared causal graph and the distinct intervention targets. During the \textbf{meta-training phase}, we are given a collection of inference tasks $t = 1, \dots, T$, associated with \textit{meta-training datasets} $\mathcal{D}_\text{train} = \{D_1, \dots, D_T\}$. Each dataset $D_t$ is a set of $N_t$ i.i.d. observations of $\biX = [X_1, \dots, X_d]^{\top}$, given as $D_t = \{\bix_n^t\}_{n=1}^{N_t} \overset{i.i.d.}{\sim} \pr^{(t)}(\biX)$, where $\bix_n^t \in \mathbb{R}^d$ ($n \in \{1, \dots, N_t\}$) is an observation drawn from joint distribution $\pr^{(t)}(\biX)$, obtained under a (hard or soft) intervention on distinct variables in $\biX$.
During the \textbf{meta-test phase}, we are given a new inference task $t'$ with a \textit{meta-test dataset} $D_{t'}'$, whose sample size can be substantially smaller than those of meta-training datasets $\mathcal{D}_\text{train}$.


Assigning samples to the task indices $t = 1, \dots, T$ requires domain knowledge about which data instances share identical, but unknown, intervention targets. We assume such domain knowledge; for example, in Example 2 in Section 1, patients administered the same pharmacological treatment are grouped into the same task. However, we do not assume knowledge of the intervention targets themselves. Moreover, test-time domains do not have to match training domains; their intervention targets may coincide with, overlap with, or differ from those seen during training.

Unlike the meta-learning setups that focus solely on observational data \citep{wu2024_biomedical_meta_causal_learning,dhir2025_metalearning_bayesian_causal_discovery}, we consider two inference targets. One is causal graph $\cg$ shared across all inference tasks $t=1,\ldots,T$, whose structure is represented by DAG adjacency matrix $\biA \in \{0, 1\}^{d \times d}$, which takes $A_{i, j} = 1$ if $X_i \rightarrow X_j$ in $\cg$; otherwise, $A_{i, j} = 0$. The other is the binary intervention target vector $\bim_t \in \{0, 1\}^d$, which takes $m_{t, i} = 1$ if and only if the $i$-th variable $X_i$ in $\biX$ is intervened on; otherwise, $m_{t, i} = 0$. 


To quantify the inference uncertainty of both $\biA$ and $\{\bim_t\}$, we aim to learn the parameters of variational distributions that approximate their posterior distributions.

\subsection{Variational Inference via Meta-Learning} \label{subsec:effective-meta-learning}

Our objective is to infer the joint posterior distribution over the shared graph $\biA$ and distinct intervention targets $\{\bim_t\}$ from the full collection of datasets $\mathcal{D}_{\mathcal{T}} = \{\mathcal{D}_\text{train}, D'_{t'}\}$ in tasks $\mathcal{T} = \{1, \dots, T, t'\}$:
\begin{equation}
\pr(\biA, \{\bim_{t}\}_{t \in \mathcal{T}} |  \mathcal{D}_{\mathcal{T}}) 
= \pr(\{\bim_{t}\}_{t \in \mathcal{T}} | \biA, \mathcal{D}_{\mathcal{T}}) \pr(\biA | \mathcal{D}_{\mathcal{T}}).
\label{eq:posterior}
\end{equation}
Unfortunately, this posterior is intractable in general, as the integral for computing the marginal likelihood $\pr(\biA|\mathcal{D}_\mathcal{T})$ cannot analytically be computed without making restrictive assumptions (e.g., linear Gaussian ANMs).

For this reason, we perform posterior approximation using a variational distribution $\pr_{\Upsilon, \Phi}$, parameterized by \textit{task-specific parameters} $\Upsilon$ and \textit{task-shared parameters} $\Phi$, where $\Upsilon$ represent the components specific to each task (e.g., the intervention targets), and $\Phi$ capture the components shared across all tasks (e.g., the causal graph structure).

Using these variational parameters, we approximate the posterior in \cref{eq:posterior} as the variational distribution factorized over meta-training and meta-test tasks $\mathcal{T}=\{1,\ldots,T, t'\}$:
\begin{align}
    \begin{aligned}
    &\pr(\biA, \{\bim_{t}\}_{t \in \mathcal{T}} \mid \mathcal{D}_{\mathcal{T}}) \\
    \approx & \pr_{\Upsilon,\Phi}(\biA \mid \mathcal{D}_{\mathcal{T}}) \prod_{t \in \mathcal{T}} \pr_{\Upsilon, \Phi}(\bim_{t} \mid \biA, D_{t}). \label{eq:variational-objective}
    \end{aligned}
\end{align}

This factorization reflects two modeling assumptions. First, we assume that intervention targets for two distinct tasks $a, b \in \mathcal{T}$ ($a \neq b$) are conditionally independent given the task-shared parameters $\Phi$: $\bim_{a} \indep \bim_{b} \mid \Phi$. Second, we assume that the shared graph $\biA$ and task-shared parameters $\Phi$ act as sufficient statistics that distill all generalizable knowledge from the other tasks. This leads to the conditional independence relation: $\bim_{t} \indep \mathcal{D}_{\mathcal{T}} \setminus D_t \mid \biA, \Phi$ for any task $t \in \mathcal{T}$. These assumptions justify the factorization over $t \in \mathcal{T}$ in \cref{eq:variational-objective}, enabling us to treat the identification of $\bim_t$ as an independent task. Note that while we index all distributions by $\Upsilon$ and $\Phi$ for notational simplicity, $\Upsilon$ is empty for the shared graph $\biA$.



\begin{remark}[\textbf{Identifiability under infinite sample regime}] \label{rem:identifiability}
Since the posterior is dominated by likelihood in this regime, we can directly apply the existing well-established identifiability results for causal discovery from multiple environments (e.g., \citet{brouillard2020_diff_interv}) to show that both $\biA$ and $\bim_{t}$ are \textbf{identifiable}.
If functional assumptions like ANMs do not hold, the causal graph can be identified only up to the I-MEC \citep{tian_and_pearl2001_IMEC}; we empirically investigate the performance of our method in such cases (\cref{appendix-linear-gaussian}).
\end{remark}

\begin{remark}[\textbf{Difficulty of finite-sample guarantees}] \label{rem:finite_sample_difficulty}
Due to the absence of meta-learning theory on its nonconvex, non-smooth, and bi-level optimization problem, establishing the consistency of finite-sample estimates (e.g., the estimated causal graph structure) is highly non-trivial. Given that recent neural-network-based approaches also lack such finite-sample guarantees \citep{brouillard2020_diff_interv,bacadi_hagele_2023}, we regard this as out of scope for this paper.
\end{remark}

\section{Proposed Method}
\label{sec-method}

This section presents \textbf{MetaCaDI}, a meta-learning framework for 
causal discovery from multiple environments.

\subsection{Variational Model Components} \label{subsec:model-components}

MetaCaDI learns the variational distributions
over shared causal DAG adjacency matrix $\biA$ and distinct intervention targets $\bim_{t}$ in task $t \in \mathcal{T}$.
Below we present the three model components: (i) the causal DAG distribution, (ii) the intervention target distribution, and (iii) the likelihood model.

\subsubsection{Differentiable Causal DAG Sampler}
\label{sec:diffdag}
We formulate the variational distribution, $\pr_{\Phi}(\biA \mid \mathcal{D}_{\mathcal{T}})$ in \cref{eq:variational-objective}, using the Differentiable Probabilistic DAG (DP-DAG) sampler \citep{diffdag_charpentier2022}, which is a DAG sampler whose parameters can be learned via a gradient-based optimization, despite the discrete nature of DAGs.

To enable gradient-based learning, the DP-DAG sampler employs a continuous relaxation of the DAG adjacency matrix. By decomposing the DAG adjacency matrix $\biA$ using a permutation matrix $\biPi \in \{0, 1\}^{d \times d}$ and a strictly upper-triangular matrix $\biU \in \{0, 1\}^{d \times d}$ as $\biA=\biPi^\top \biU \biPi$, it approximately computes the gradient using continuous relaxations $\tilde{\biU} \in \R^{d \times d}$ and $\tilde{\biPi} \in \R^{d \times d}$,
each of which can be sampled using the Gumbel-Softmax distribution \citep{Jang2016_gumbel, Maddison2016_gumbel} and the Gumbel-Top-$k$ trick \citep{kool2019_gumbel_topk}, respectively.

The resulting causal DAG distribution model
$\pr_{\Phi}(\biA | \mathcal{D}_\mathcal{T})$ only contains the parameters of these distributions, 
denoted by $\boldsymbol{\phi}$ and $\boldsymbol{\psi}$. Since causal graph $\biA$ is shared across all tasks, we treat them as task-shared parameters $\Phi$.

\subsubsection{Differentiable Intervention Target Identifier}
\label{sec:differentiable_intervention}
Analogous to the DAG distribution model in \cref{sec:diffdag}, we model the intervention target distribution $\pr_{\Upsilon, \Phi}(\bim_{t} \mid \biA, D_{t})$ by designing a differentiable sampler.

Since the Bernoulli sampling operation over the binary vector $\bim_{t} \in \{0, 1\}^d$ is non-differentiable, we use a sampler of its continuous relaxation $\tilde{\bim}_{t} \in \mathbb{R}^d$, formulated as a Gumbel-Softmax distribution with logit parameters $\boldsymbol{\zeta}^{t} \in \mathbb{R}^d$.

However, learning this logit vector only from dataset $D_t$ is challenging, as intervention target inference ideally requires comparison between interventional and observational data distributions, which is infeasible in our setting. As we will describe in \Cref{subsec:modeling},
we address this challenge not by comparing datasets
but by comparing the \textit{predicted} values of the variables with and without interventions. We obtain these predicted values using the likelihood model.

\subsubsection{Likelihood Model}

To learn the parameters of the variational distributions defined in \cref{eq:variational-objective}, we optimize the Evidence Lower Bound (ELBO) by using the sampled $\biA$ and $\bim_t$, to model the likelihood $\pr_{\Upsilon, \Phi}(D_t \mid \biA, \bim_t)$. To enable efficient task adaptation, we instantiate the model with ANM; however, other models can be used in our general meta-learning framework.


To model the likelihood, we represent the parents $\text{PA}(X_i)$ of each variable $X_i \in \biX$ by masking the input variable vector $\biX$ with $\bitA_i \in \{0, 1\}^d$, obtained from the $i$-th column of the DAG adjacency matrix $\biA$. Using the sampled binary intervention indicator $m_{t,i} \in \{0,1\}$ as a switch between the observational and interventional mechanisms, we formulate the structural equation of each $X_i$ as
\begin{align}
X_i = (1 - m_{t,i})& \cdot f_i(\bitA_i  \circ \biX; \boldsymbol{\Theta}_\text{obs}) \nonumber \\ & + m_{t,i} \cdot f^{\text{I}}_i(\bitA_i \circ \biX; \boldsymbol{\Theta}_\text{int}) + \epsilon_i,
\label{eq:scm_main}
\end{align}
where $\epsilon_i \sim \mathcal{N}(0, 1)$ is standard Gaussian noise, $\circ$ denotes the Hadamard product, and $f_i$ and $f_i^\text{I}$ are functions representing the observational and interventional mechanisms, respectively. We parameterize these functions as multi-layer perceptrons (MLPs) with parameters $\boldsymbol{\Theta}_\text{obs}$ and $\boldsymbol{\Theta}_\text{int}$; we describe how we split them into $\Upsilon$ and $\Phi$ in \cref{subsec:closed-form-solver}.

By parameterizing $f^\text{I}_i$ as a flexible function and using the binary indicator $m_{t,i}$ in \cref{eq:scm_main}, MetaCaDI inherently accommodates hard, soft, and complex mixtures of interventions across variables without requiring model architecture changes.


\subsection{Modeling Strategies for Efficient Task Adaptation} \label{subsec:modeling}

Directly sampling from the variational model (\cref{subsec:model-components}) to infer the graph and interventions is highly challenging, as it requires estimating a massive number of parameters from a few-shot target dataset. To address this, the following subsections introduce our feature construction (\cref{subsubsec:logit_parameters}) and analytical adaptation (\cref{subsec:closed-form-solver}) methods, which efficiently reduce the parameter space.

\subsubsection{Towards Efficient Logit Parameter Learning} 
\label{subsubsec:logit_parameters}

The first challenge in few-shot adaptation in our setup is to train the logit parameters $\boldsymbol{\zeta}^t$ of the intervention target distribution, under the infeasibility of comparing interventional and observational data distributions (\cref{sec:differentiable_intervention}).

Our strategy for this challenge is twofold. First, we extract a feature vector $\biF_t$ that contains the differences between the predicted variable values obtained by applying the observational and interventional mechanisms in \eqref{eq:scm_main} to the support set $D_t^s$. Then we obtain the logit parameter values by 
\begin{equation}
    \boldsymbol{\zeta}^t = g_{\boldsymbol{\eta}}(\biF_t), \label{eq:logit_predictor}
\end{equation}
where $g_{\boldsymbol{\eta}}$ is a neural network with parameters $\boldsymbol{\eta}$.

This modeling strategy has two advantages.
First, by leveraging the dependence of input feature $\biF_t$ on distinct task $t$,
we can treat the parameters $\boldsymbol{\eta}$ as task-shared parameters, thus reducing the number of task-specific parameters $\Upsilon$, which need to be trained on small support sets.
Second, it can effectively specify informative features for identifying the intervention target $\bim_t$
by including them in the input feature $\biF_t$.

\paragraph{Detailed feature design.} We obtain feature vector $\biF_t$ by taking three steps. First, we make predictions by applying the observational and interventional mechanisms in \cref{eq:scm_main} to each data instance $\bix_n^s$ in support set $D_t^s$:
\begin{align*}
        [\hat{\bix}^s_{\text{obs}, n}]_i = f_i(\bitA_i \circ \bix_n^s; \boldsymbol{\Theta}_\text{obs})\ \text{and} \  
        [\hat{\bix}^s_{\text{int}, n}]_i = f^\text{I}_i(\bitA_i \circ \bix_n^s; \boldsymbol{\Theta}_\text{int}).
\end{align*}
By concatenating these vectors in a row-wise manner,
we obtain the two $N_t^s \times d$ matrices 
$\hat{\mathbf{X}}^s_\text{obs}$ and $\hat{\mathbf{X}}^s_\text{int}$, whose $n$-th row vectors are $\hat{\bix}^s_{\text{obs}, n}$ and $\hat{\bix}^s_{\text{int}, n}$, respectively.

Second, using the matrices $\hat{\mathbf{X}}^s_\text{obs}$ and $\hat{\mathbf{X}}^s_\text{int}$, together with the support set data matrix $\mathbf{X}^s_t \in \mathbb{R}^{N^s_t \times d}$, we engineer a feature matrix, $\mathbf{C}_t \in \mathbb{R}^{N_t^s \times 9d}$ by concatenating the following nine $N^s_t \times d$ matrices in a column-wise manner:
\begin{align*}
    \mathbf{C}_t = \big[ & \mathbf{X}^s_t, \hat{\mathbf{X}}^s_\text{obs}, \mathbf{X}^s_t - \hat{\mathbf{X}}^s_\text{obs}, (\mathbf{X}^s_t - \hat{\mathbf{X}}^s_\text{obs})^2, \\
    & \hat{\mathbf{X}}^s_\text{int}, \mathbf{X}^s_t - \hat{\mathbf{X}}^s_\text{int}, (\mathbf{X}^s_t - \hat{\mathbf{X}}^s_\text{int})^2, \boldsymbol{\mu}(\mathbf{X}^s_t), \boldsymbol{\sigma}(\mathbf{X}^s_t) \big],
\end{align*}
where $\boldsymbol{\mu}(\mathbf{X}^s_t)$ and $\boldsymbol{\sigma}(\mathbf{X}^s_t)$ are the broadcasted column-wise mean and standard deviation of $\mathbf{X}^s_t$, respectively.
This design explicitly compares the interventional and observational distributions through their residuals, i.e., the difference between observed data and SCM predictions.

Finally, we flatten the matrix $\mathbf{C}_t$ 
into a feature vector $\biF_t$ 
so that the values of $\biF_t$ are invariant to the order of data instances in $D_t^s$.
To do so, 
we formulate the two permutation-invariant pooling layers
based on the Deep Sets architecture \citep{zaheer2017deepsets}.
Using an MLP $h_{\boldsymbol{\varsigma}}: \mathbb{R}^{9d} \to \mathbb{R}^k$ with task-shared parameters $\boldsymbol{\varsigma}$, 
these layers compute the \textbf{mean of embeddings} and the \textbf{embedding of the mean}:
\begin{align*}
    \mathbf{z}_{\text{ME}} = \frac{1}{N_t^s} \sum_{n=1}^{N_t^s} h_{\boldsymbol{\varsigma}}(\bic_{t,n})\ \text{and}\  
    \mathbf{z}_{\text{EM}} = h_{\boldsymbol{\varsigma}}\left(\frac{1}{N_t^s} \sum_{n=1}^{N_t^s} \bic_{t,n}\right),    
\end{align*}
where embedding $\bic_{t,n}$ is the $n$-th row of $\mathbf{C}_t$.
Using $\mathbf{z}_{\text{ME}}$ and $\mathbf{z}_{\text{EM}}$, we obtain $\biF_t$ as a vector of size $2k$:
\begin{equation*}
    \biF_t = [\mathbf{z}_{\text{ME}}^\top, \mathbf{z}_{\text{EM}}^\top]^\top.
\end{equation*}

\begin{remark}[\textbf{Performance gain by feature design}]
Our extensive ablation studies demonstrate the strong performance gain by this feature design. The inference performance significantly degrades when removing the residual matrices in $\mathbf{C}_t$ or replacing the Deep Sets pooling with naive pooling architectures. See \cref{appendix-ablation} for details.
\end{remark}

\subsubsection{Analytical Adaptation via Closed-Form Solvers} 
\label{subsec:closed-form-solver}

The second challenge is to learn the MLP parameters $\boldsymbol{\Theta}_{\text{int}}$ for the interventional mechanisms $f^{\text{I}}_i$ in \cref{eq:scm_main}, which are specific to each task $t$ but whose number of parameters is typically large in our few-shot settings. 

Our key idea to address this challenge is to reduce the number of task-specific parameters $\Upsilon$ by treating only the last MLP layer of $f^{\text{I}}_i$, denoted by $\biw_i$, as task-specific parameters:
\begin{equation}
f^{\text{I}}_i(\bitA_i \circ \biX; \boldsymbol{\Theta}_\text{int}) = \biw_i^\top h_{\boldsymbol{\Theta}'_\text{int}}(\bitA_i \circ \biX), \label{eq:int_MLP}
\end{equation}
where $h_{\boldsymbol{\Theta}'_\text{int}}$ is the preceding layers parameterized by task-shared parameters $\boldsymbol{\Theta}'_\text{int} = \boldsymbol{\Theta}_\text{int} \setminus \biw_i$.
Crucially, this linear adaptation is applied strictly to the highly non-linear outputs of the feature extractor $h_{\boldsymbol{\Theta}'_\text{int}}$, preserving the expressivity to capture complex, non-linear relationships while leveraging the robust linear solver to prevent overfitting on scarce data.

Under this formulation, we can analytically obtain the optimal weights $\hat{\biw}^{(t)}_i$ best fitted to the support set $D^s_t$ by solving a least-squares problem with an $\ell_2$ regularizer (i.e., Ridge regression):
\begin{equation}
\min_{\biw_i} \left\| (\mathbf{X}^s_t)_{:,i} - \mathbf{H}_i \biw_i \right\|_2^2 + \lambda \left\| \biw_i \right\|_2^2, \label{eq:ridge}
\end{equation}
where $(\mathbf{X}^s_t)_{:,i}$ is the $i$-th column of the support set data matrix $\mathbf{X}^s_t$, 
$\mathbf{H}_i \in \mathbb{R}^{N^s_t \times d_h}$ is the hidden feature matrix extracted by applying the preceding layers of the MLP $h_{\boldsymbol{\Theta}'_\text{int}}$ to each support set instance, and $\lambda \geq 0$ is a regularization parameter. The regularized least-squares problem in \cref{eq:ridge} has a well-known closed-form solution:
\begin{equation}
\label{eq:closed-form-solution}
\hat{\biw}^{(t)}_i = \left( \mathbf{H}_i^\top \mathbf{H}_i + \lambda \mathbf{I} \right)^{-1} \mathbf{H}_i^\top (\mathbf{X}^s_t)_{:,i},
\end{equation}
where $\mathbf{I}$ is the identity matrix.

As discussed in existing work on meta-learning~\citep{snell2017_prototypical_networks, Bertinetto2018MetalearningWD}, such analytical task-specific parameter optimization enables fast and data-efficient task adaptation on $D^s_t$.
While general-purpose meta-learning methods (e.g., MAML \citep{maml_finn2017}) offer high expressivity via an iterative gradient-based optimization, they are prone to overfitting and instability when adapting to extremely small support sets (e.g., $N^s_t = 10$).
Our analytical approach achieves good balance between expressivity (via the shared feature extractor) and robust estimation (via the closed-form solver), leading to superior empirical performance, as we will demonstrate in \cref{sec:experiments}.

\subsection{Training Objective} \label{subsec:training_objective}
Prior to training, we randomly partition each meta-training dataset $D_t$ into a fixed \textbf{support set} $D_t^s$ and a \textbf{query set} $D_t^q$, which are illustrated in \cref{fig:prob-setup} as the two disjoint sets with green and red data instances, respectively. 
The small support set mimics the few-shot setting, while the query set evaluates task adaptation via a loss function.

During the meta-training phase, 
MetaCaDI takes two steps for each meta-training task $t$.
First, it fits task-specific parameters $\Upsilon = \{\biw_1, \dots, \biw_d\}$ to the support set $D^s_t$ by leveraging the closed-form solution in \cref{eq:closed-form-solution}.
Then, with the trained $\Upsilon$, it performs gradient-based updates for the task-shared parameters 
$\Phi = \{
    \biphi,
    \boldsymbol{\psi}, \boldsymbol{\eta}, \boldsymbol{\varsigma}, \boldsymbol{\Theta}_\text{obs}, \boldsymbol{\Theta}_\text{int}'
\}$ to minimize an expected loss over all meta-training tasks $t = 1,\dots,T$.

To approximate the posterior by maximizing the ELBO, we minimize the expected loss
\begin{equation*}
    	\underset{\Phi}{\text{min}} \E_{t} [\E_{\biA, \bim_t} [- \log \pr(D^q_t \mid \biA, \bim_t)] + \Omega_{\Phi, t}],
\end{equation*}
where $\log \pr(D^q_t \mid \biA, \bim_t)$ is the log-likelihood over the query set $D^q_t$ for task $t$, and $\Omega_{\Phi, t}$ is a regularization term.

We estimate this expected loss as an empirical average over meta-training tasks $t= 1, \dots, T$.
By approximating the expectation over $\biA$ and $\bim_t$ using a single sample from the differentiable samplers in \cref{subsec:model-components}, we formulate our meta-training objective as
\begin{align}
 \mathcal{L}(\Phi) = \frac{1}{T}\sum_{t=1}^T (\mathcal{L}_{\text{R},t} + \lambda_{I}\mathcal{L}_{\text{I},t} + \lambda_{H}\mathcal{L}_{\text{H},t} ) + \lambda_{G}\mathcal{L}_{G}, \label{eq:meta-training-objective}
\end{align}
which is the sum of the following four loss components weighted by regularization parameters $\lambda_I, \lambda_H, \lambda_G \geq 0$:
\begin{itemize}[noitemsep, topsep=0pt]
    \item \textbf{Reconstruction Loss} $\mathcal{L}_{\text{R},t}$ which uses the mean squared error (MSE) over the query set $D^q_t$;
    \item \textbf{Intervention sparsity regularizer} $\mathcal{L}_{\text{I},t}$, which applies an $l^1$ regularizer to logit vector $\boldsymbol{\zeta}^t$ to enforce sparsity on the intervention target vector $\bim_t$;
    \item \textbf{Residual independence regularizer} $\mathcal{L}_{\text{H},t}$, which penalizes the dependence of residuals using the Hilbert-Schmidt Independence Criterion (HSIC) \citep{gretton2007_hsic} to enforce the causal sufficiency assumption;
    \item \textbf{Graph sparsity regularizer} $\mathcal{L}_{G}$, which encourages sparsity on the edge presence probability in the causal graph using a Kullback-Leibler (KL) divergence.
\end{itemize}

We detail the formulation of each component in \cref{appendix-training-objective}.

\paragraph{Meta-test adaptation.} During the meta-test phase, we fine-tune the task-specific parameters $\Upsilon$ on the meta-test dataset $D'_{t'}$ using the closed-form solution in \cref{eq:closed-form-solution}. While the task-shared parameters $\Phi$ could theoretically be fine-tuned as well, we hypothesize that the severe scarcity of the meta-test data would make such updates highly prone to overfitting. Therefore, we keep $\Phi$ frozen during adaptation. We empirically validate this design choice in \cref{sec:experiments} by including variants (i.e., ablations) of our method that update these task-shared parameters.

\section{Experiments}
\label{sec:experiments}

\subsection{Experimental Setup}

\paragraph{Baselines.} We compare MetaCaDI against four baselines: the constraint-based \textbf{JCI-PC} \citep{jci_mooij_jmlr2020}, the score-based \textbf{UT-IGSP} \citep{jci_extension_squires2020}, the continuous optimization-based \textbf{DCDI-G} \citep{brouillard2020_diff_interv}, and the multitask Bayesian framework \textbf{BaCaDI} \citep{bacadi_hagele_2023}.
To validate the effectiveness of our analytical adaptation (\cref{eq:closed-form-solution}), we compare against three variants that employ gradient-based MAML updates \citep{maml_finn2017}: \textbf{MetaCaDI-I-MAML}, which updates only the interventional mechanism $f^\text{I}_i$; \textbf{MetaCaDI-IO-MAML}, which updates both the observational and interventional likelihood parameters; and \textbf{MetaCaDI-Full-MAML}, which updates all model parameters including those of the causal graph. We refer to our proposed method as \textbf{MetaCaDI-Analytical}.


\begin{figure*}[t]
\centering
\includegraphics[width=0.9\textwidth]{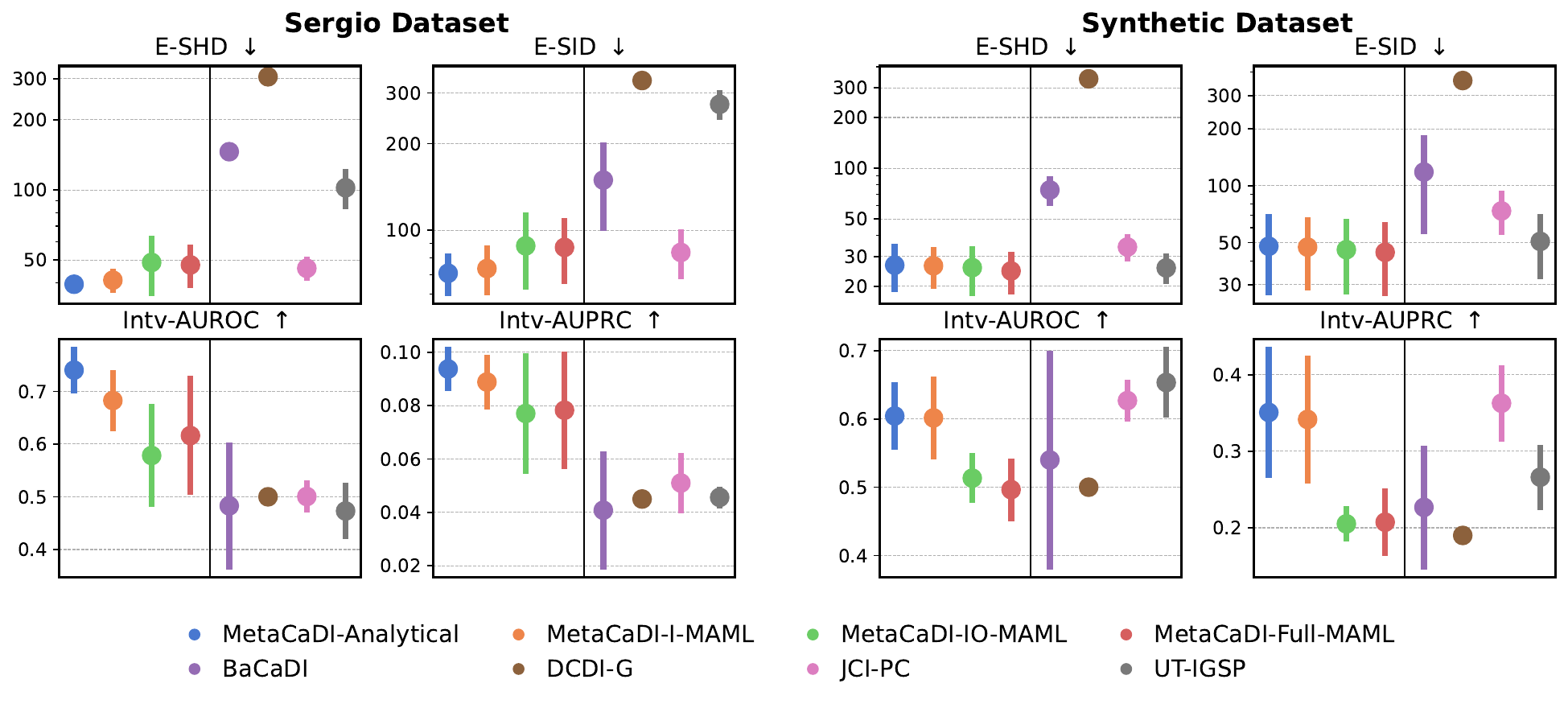}
\caption{Performance comparison on the SERGIO and synthetic datasets. Results are averaged over 20 independent simulations per method, with error bars indicating the standard deviation. Four methods on the left of the black vertical line are based on our proposal: our method and its three variants. $\downarrow$ and $\uparrow$ indicate "lower is better" and "higher is better", respectively.}
\label{fig:main_results}
\end{figure*}

\paragraph{Data.}
We use synthetic and semi-synthetic datasets, as a rigorous evaluation is \textbf{infeasible on real-world datasets} due to the absence of well-established benchmarks with ground-truth causal graphs and intervention targets.

\textbf{Synthetic} data is generated using non-linear Gaussian-Process-based ANMs with soft interventions. Semi-synthetic data is sampled from \textbf{SERGIO} \citep{sergio2020}, a \textit{realistic} simulator with parameters tuned to real gene expression data to capture complex nonlinearity and biological noise,  with hard interventions. 



The ground truth causal graphs for the synthetic and semi-synthetic datasets are randomly sampled from the Erdős-Rényi (ER) model and the scale-free model, respectively. In our main experiments, we set the number of variables to $d=20$ for both datasets. We generate $T=20$ training datasets with size $N_t=110$ and $T'=20$ test datasets with size $N_{t'}'=10$. For MetaCaDI, we split each (meta-)training dataset into a support set of size $N^s_t = 10$ and a query set of size $N^q_t = 100$. See \cref{appendix-data-gen} for details.

\paragraph{Performance Evaluation.}
For a fair comparison, we evaluate the test performance by enforcing a strict \textit{single-task inference} protocol: All methods use only the single (meta-)test dataset (of size $N_{t'}'=10$) for the specific test task being evaluated, without access to the full pool of test tasks (see \cref{appendix-baseline-adaptation-protocol} for the detailed procedure for baselines).

For graph discovery, we use the \textbf{Expected Structural Hamming Distance (E-SHD)} and the \textbf{Expected Structural Intervention Distance (E-SID)} \citep{peters2013_sid}. 
For intervention target identification, we use the \textbf{Interventional AUROC (Intv-AUROC)} and \textbf{Interventional AUPRC (Intv-AUPRC)}. See \cref{appendix-exp-setup} for details. 

\subsection{Results}

We compute the mean and standard deviation of each performance metric over 20 independent simulations. \cref{fig:main_results} shows the results on the SERGIO and synthetic datasets. Below we discuss them for each front of evaluation. 

\paragraph{Intervention Target Identification.}
Despite the challenging setting with only $N'_{t'}=10$ observations, our MetaCaDI-Analytical significantly outperforms all the four existing methods on SERGIO datasets, and achieves better or competitive performance on synthetic datasets, demonstrating its superior performance on such complex datasets as SERGIO.

On the SERGIO datasets, our method achieves the best mean Intv-AUROC of $0.740$ and mean Intv-AUPRC of $0.094$, whereas all of BaCaDI, DCDI-G, JCI-PC, and UT-IGSP struggle to exceed random chance ($\sim0.50$) in this challenging setting ($p < 0.001$). These results suggest that existing methods cannot capture the complex, realistic intervention mechanisms from the limited data, thus highlighting the importance of our meta-learning approach that enables the rapid adaptation to such scarce datasets.

\paragraph{Causal Graph Discovery.}
Compared to the existing methods, our MetaCaDI-Analytical achieves the best performance in the SERGIO datasets and proves to be better or highly competitive on the synthetic datasets, demonstrating its successful performance for extracting the shared causal graph knowledge across datasets. 

On both datasets, DCDI-G and BaCaDI exhibit much poorer performance than our MetaCaDI-Analytical, despite all three methods being based on continuous optimization. A possible reason for this performance gap is that DCDI-G and BaCaDI cannot search over the space of \textit{valid} DAGs, as they rely on a regularization term that only \textit{encourages} acyclicity but does not strictly enforce it. By leveraging the DP-DAG sampler formulation, our MetaCaDI-Analytical effectively infers the posterior distribution over valid DAGs, thus yielding remarkable performance in graph discovery.

\paragraph{Ablation on Analytical Adaptation.}
To validate the benefit of our analytical task adaptation strategy (\cref{subsec:closed-form-solver}), we compare our MetaCaDI-Analytical against its three variants (MetaCaDI-I-MAML, MetaCaDI-IO-MAML, and MetaCaDI-Full-MAML), based on gradient-based updates. 

\begin{figure}[t]
    \centering
    \includegraphics[width=\linewidth]{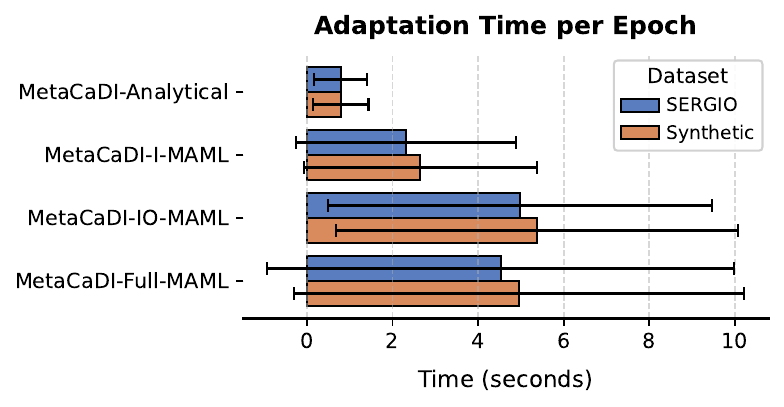}
    \caption{Runtime comparison with MetaCaDI variants}
    \label{fig:adaptation_per_epoch}
\end{figure}

In terms of intervention target identification (\cref{fig:main_results} bottom), our MetaCaDI-Analytical outperforms all variants in 9 of 12 comparisons (3 variants $\times$ 2 datasets $\times$ 2 metrics), with statistical significance ($p < 0.05$). 
Moreover, as shown in the runtime comparison in \cref{fig:adaptation_per_epoch}, it achieves
approximately 5.7 times faster task adaptation than MetaCaDI-Full-MAML on average for the SERGIO datasets. This 5.7x faster adaptation directly addresses the real-time requirements of \textbf{Example 1}.
Moreover, these results underscore the effectiveness and efficiency of our analytical adaptation, enabling real-time inference from limited data. 
More detailed runtime comparisons can be found in \cref{appendix-runtime-comparison}.

\subsection{Additional Experiments}

We further test our MetaCaDI with additional experiments. We summarize the key findings, with details in \cref{appendix-additional-experiments}.

\begin{figure}[t]
\centering
\includegraphics[width=0.9\linewidth]{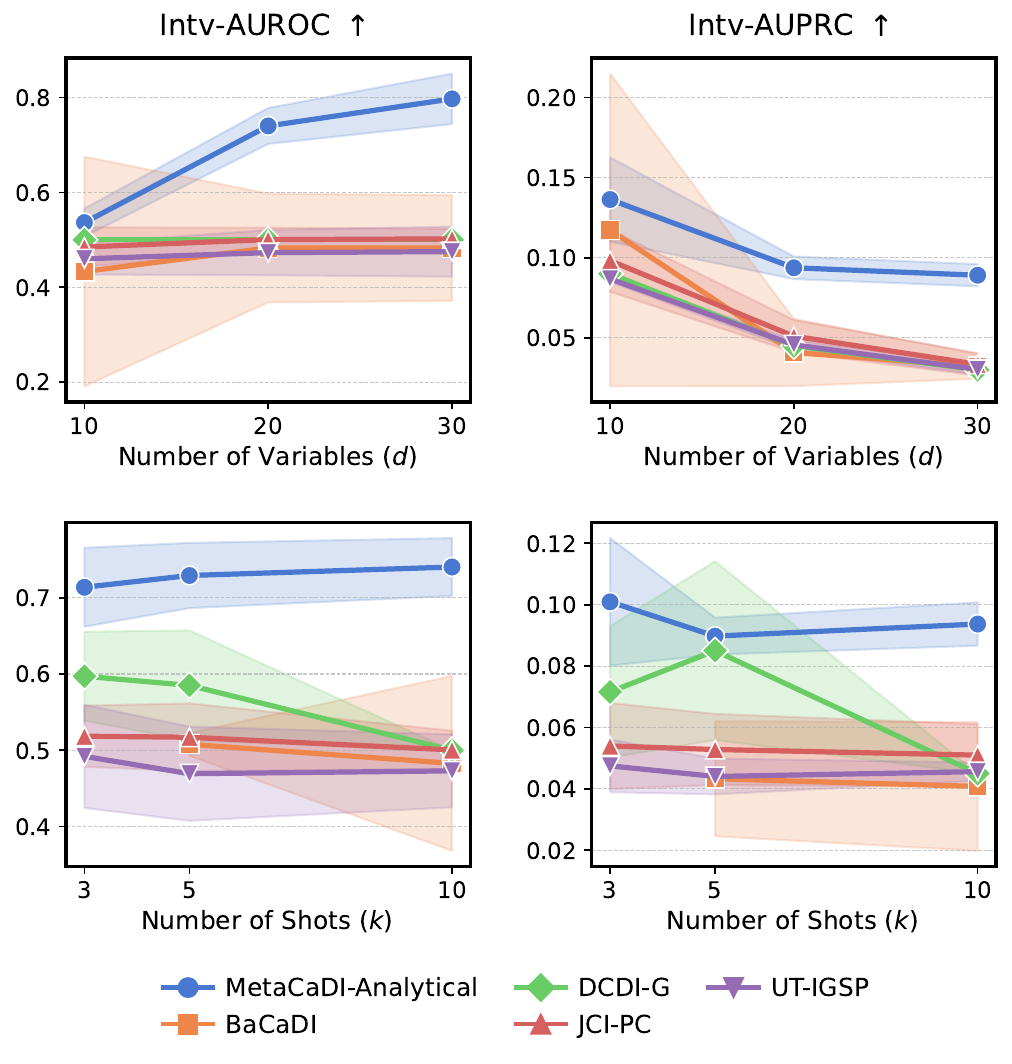}
\caption{Performance sensitivity to the number of variables ($d$) and the support set size ($k$) on the SERGIO datasets.}
\label{fig:combined_sensitivity_sergio}
\end{figure}


\paragraph{Computational Complexity and Scalability.}
We evaluate scalability by varying variables up to $d=100$. MetaCaDI consistently outperforms baselines, with its efficiency advantage widening substantially at scale; while methods like BaCaDI and JCI-PC often timeout past 50 hours at high dimensions, MetaCaDI maintains superior performance and offers an order-of-magnitude overall speedup ($\sim14.5$ hours vs. $\sim245$ hours for BaCaDI on full pipelines). See \cref{appendix-sensitivity-d}.

\paragraph{Robustness to Distribution Shifts.}
To test the robustness of MetaCaDI, we introduced two types of perturbations exclusively during meta-test data generation: (1) noise variance shifts, where the exogenous noise of each variable is scaled by $\sigma \sim \text{Uniform}(0.5, 3.0)$, and (2) causal mechanism drift, where the structural functions are scaled by $\delta \sim \text{Uniform}(0.8, 1.2)$. MetaCaDI robustly maintained its structural recovery and intervention identification performance under both perturbation types. Detailed results and the formal data generation process are provided in \cref{appendix-distribution-shifts}.

\paragraph{Few-Shot Robustness.}
We test MetaCaDI in the ultra-low data regime with the (meta-)test dataset size $k \in \{3, 5, 10\}$ by setting the support set size to $k$. 
In these tests, we employ a similar setup as the main experiments: using 20 meta-training tasks and 20 meta-test tasks and matching the support set size to $k$ while fixing the query set size to $100$.
\cref{fig:combined_sensitivity_sergio} shows the results on the SERGIO datasets. Again, MetaCaDI outperforms all baselines. The performance gap is significant even in the ultra-low data regime ($k=3$), where most baselines degrade to random chance (Intv-AUROC $\approx0.5$) or fail to adapt entirely (notably, BaCaDI).
This high robustness on as few as 3 samples directly addresses the data-scarce medical challenges motivated in \textbf{Example 2}, highlighting the significance of our meta-learning framework for enabling robust inference under severe data scarcity, which is common in real-world settings. 
Detailed results and analysis are provided in \cref{appendix-sensitivity-k}.

\paragraph{Ablation on Feature Design.} 
Our MetaCaDI consistently outperforms the variants that do not use the designed features for logit parameter learning (\cref{subsubsec:logit_parameters}). See \cref{appendix-ablation} for details.

\section{Conclusion}
\label{sec:discussion}
To tackle the challenging few-shot scenarios in causal discovery from multiple environments, we introduced MetaCaDI, the first framework to cast the joint inference of causal graphs and unknown interventions as a meta-learning problem. By leveraging a closed-form analytical adaptation, our method bypasses the instability of standard meta-learning approaches, enabling rapid and robust adaptation to the few-shot scenarios typical of data-scarce real-world scenarios. Empirical results confirm that MetaCaDI significantly outperforms state-of-the-art baselines in both structure recovery and intervention identification, offering a practical solution for unraveling complex systems under severe data scarcity.


\bibliography{refs}

\newpage

\onecolumn

\title{MetaCaDI: A Meta-Learning Framework for Causal Discovery\\from Multiple Environments with Unknown Interventions\\(Supplementary Material)}
\maketitle

\appendix
\crefalias{section}{appendix}
\crefalias{subsection}{appendix}
\crefalias{subsubsection}{appendix}
\section{Baselines, Datasets, and Metrics}
\label{appendix-exp-setup}

This section presents additional details 
regarding the baselines, data generation processes, and the metrics for performance evaluation.

\subsection{Baselines}
We compare our proposed method, \textbf{MetaCaDI-Analytical}, against seven baselines. Four are existing methods designed for causal discovery from multiple environments with unknown interventions:
\begin{itemize}
\item Joint Causal Inference based on the Peter-Clark (PC) algorithm (\textbf{JCI-PC}) \citep{jci_mooij_jmlr2020}, a constraint-based method that performs conditional independence tests involving auxiliary \textit{context variables} representing the intervention targets.
\item Unknown-Target Interventional Greedy Sparsest Permutation (\textbf{UT-IGSP}) \citep{jci_extension_squires2020}, a score-based method that solves a discrete optimization problem using a greedy search, guided by a score function that evaluates the fit of a causal graph to the interventional datasets.
\item Differentiable Causal Discovery from Interventional Data (\textbf{DCDI-G}) \citep{brouillard2020_diff_interv}, a continuous-optimization method relying on a differentiable regularizer that encourages the acyclicity of the learned graph structure.
\item Bayesian Causal Discovery with Unknown Interventions (\textbf{BaCaDI}) \citep{bacadi_hagele_2023}, a multitask Bayesian framework that learns a posterior distribution over the causal graph structure and intervention targets, leveraging the shared information across multiple tasks.
\end{itemize}
The other three are variants of our method, which do not use our proposed analytical, closed-form solvers for task adaptation. In particular, we consider the following variants:
\begin{itemize}
    \item \textbf{MetaCaDI-I-MAML}, which learns the parameters of the interventional mechanisms in our likelihood model (i.e., $\boldsymbol{\Theta}_\text{int}$) by performing gradient-based updates.
    \item \textbf{MetaCaDI-IO-MAML}, which trains all likelihood model parameters (i.e., $\boldsymbol{\Theta}_\text{int}$ and $\boldsymbol{\Theta}_\text{obs}$) using gradient-based updates.
    \item \textbf{MetaCaDI-Full-MAML}, which updates all model parameters, including those governing the causal graph structure, using gradient-based updates.
\end{itemize}
For a fair and standardized comparison, the implementations of all baseline methods are taken from the official BaCaDI repository\footnote{\texttt{https://github.com/haeggee/bacadi}}.


\subsection{Datasets and Generation Processes}
\label{appendix-data-gen}
Our experiments utilize fully synthetic and semi-synthetic datasets, both of which have known ground-truth causal graphs and intervention targets, thus enabling rigorous evaluation.

The \textbf{synthetic datasets} are generated using a custom pipeline adapted from \citet{bacadi_hagele_2023}.
We randomly generate the underlying causal graphs from the Erd\H{o}s-R\'enyi (ER) graph models.
Using the generated causal graphs, we simulate data from nonlinear ANMs, where the nonlinear function is sampled from a Gaussian Process (GP) prior with a Mat\'ern kernel with parameter $\nu=0.5$,
and the additive noise is drawn from a Gaussian distribution with mean zero and standard deviation $\sigma=1.5$. 
To generate interventional datasets, we perform soft interventions, where the function drawn from the GP prior with a Mat\'ern kernel is replaced with a new function drawn from a GP prior with a Radial Basis Function (RBF) kernel.

The \textbf{semi-synthetic datasets} are simulated from the SERGIO simulator \citep{sergio2020}, which generates realistic single-cell gene expression data.
For direct comparability, we exactly follow the same simulation protocol with \citet{bacadi_hagele_2023}. This protocol involves sampling the underlying causal graphs from a scale-free model with an expected node degree of 2 (SF-2), setting the number of nodes to $d=20$, and simulating interventional datasets by performing $M=10$ hard interventions, which force the production rate of each target gene to zero.

We use these data-generation procedures to create (meta-)training and (meta-)test datasets. For our MetaCaDI-Analytical and its variants, we generate $T'=20$ datasets, each comprising only $N_{t'}'=10$ data instances to simulate the challenging few-shot inference scenario.

\subsection{Evaluation Metrics}
\label{appendix-evaluation-metrics}
To provide a comprehensive evaluation, we assess performance on two distinct tasks: causal graph discovery and intervention target identification. For metrics calculated over a posterior distribution, we report the expected value by averaging the metric over all posterior samples.

To evaluate the performance of causal graph discovery, we use four metrics:
\begin{itemize}
    \item Expected Structural Hamming Distance (\textbf{E-SHD}), which is the expected number of edge additions, deletions, or reversals needed to convert the inferred causal graph structure to the ground truth graph. Lower is better.
    \item Expected Structural Intervention Distance (\textbf{E-SID}), which evaluates the ability of the inferred causal graph structure to identify interventional distributions \citep{peters2013_sid}. In particular, this metric evaluates whether the inferred graph structure correctly specifies the valid adjustment set for the identification of interventional distributions. Roughly speaking, it takes a high value if the inferred graph structure leads to significantly incorrect causal implications. Lower is better.
\end{itemize}
Regarding intervention target identification, we use the following two classification metrics:
\begin{itemize}
    \item Interventional AUROC \textbf{(Intv-AUROC)}, which measures the AUROC score in the task of identifying true intervention targets. Higher is better.
    \item Interventional AUPRC \textbf{(Intv-AUPRC)}, which evaluates the AUPRC score in intervention target identification. As with edge AUPRC, it is better suited than Intv-AUROC for the sparse intervention setting, where only a few variables are intervened in all datasets. Higher is better.
\end{itemize}

\section{Detailed Training Objective}
\label{appendix-training-objective}
As described in \cref{subsec:training_objective},
we learn the task-shared parameters 
$\Phi = \{
    \boldsymbol{\phi}, \boldsymbol{\psi}, \boldsymbol{\eta}, \boldsymbol{\Theta}_\text{obs}, \boldsymbol{\Theta}_\text{int}'
    , \boldsymbol{\varsigma}
    \}$
by minimizing the following loss function:
\begin{align*}
 \mathcal{L}(\Phi) = \frac{1}{T}\sum_{t=1}^T (\mathcal{L}_{\text{R},t} + \lambda_{I}\mathcal{L}_{\text{I},t} + \lambda_{H}\mathcal{L}_{\text{H},t} ) + \lambda_{G}\mathcal{L}_{G}.
\end{align*}
Below we detail the four components of this loss function.
\begin{itemize}
    \item \textbf{Reconstruction Loss} $\mathcal{L}_{\text{R},t}$ is the expected negative log-likelihood of the query set $D^q_t$ for task $t$. Under the standard Gaussian additive noise assumption, such negative log-likelihood equals the Mean Squared Error (MSE):
    \begin{align}
    \mathcal{L}_{\text{R},t} &= - \log \pr(D^q_t \mid \biA, \bim_t) \\
    &= \frac{1}{d \cdot N^q_t} \sum_{i=1}^d \sum_{n=1}^{N^q_t} (x^q_{n, i} - \hat{x}^q_{n, i})^2, \label{eq:reconstruction-loss}
    \end{align}
    where $\hat{x}_{n, i}^q$ is the predicted value of variable $X_i$ for the $n$-th observation $\bix^q_n$ in the query set $D^q_t$.
    \item \textbf{Intervention Sparsity Loss} $\mathcal{L}_{\text{I},t}$ is a sparsity constraint on the predicted probabilities that each variable $X_i$ in $\biX$ is intervened in task $t$. Formally, it imposes a $\ell^1$ penalty on the probabilities derived from the logit vector $\boldsymbol{\zeta}^t$ of intervention target identification for task $t$ as 
    \begin{equation*}
    \mathcal{L}_{\text{I},t} = \| \sigma(\boldsymbol{\zeta}^t) \|_1,
    \end{equation*}
    where $\sigma(\cdot)$ is the sigmoid function.
    \item \textbf{Residual Independence Loss} $\mathcal{L}_{\text{H},t}$ is a penalty term for encouraging the independence among residuals $x^q_{n, i} - \hat{x}^q_{n, i}$ in \cref{eq:reconstruction-loss}
    to enforce the causal sufficiency assumption. To measure this independence, we use the Hilbert-Schmidt Independence Criterion (HSIC) \citep{gretton2007_hsic} for this purpose.
    \item \textbf{Graph Sparsity Loss} $\mathcal{L}_{G}$ is a KL divergence term that encourages sparsity in edge presence probabilities. Letting $\pr_{\biphi}(\biU)$ be the Gumbel-Softmax-based variational distribution over upper-triangular matrix $\biU \in \{0, 1\}^{d \times d}$ (\cref{sec:diffdag}), this KL divergence can be expressed as
    \begin{equation*}
    \mathcal{L}_{G} = \text{KL}(\pr_{\biphi}(\biU) \| \pr(\biU)),
    \end{equation*}    
    where $\pr(\biU)$ is a prior distribution, which we set to be a Bernoulli distribution with small success probabilities; see \cref{tab:sweep_config} for the setting of this success probability hyperparameter, $p_G$.
\end{itemize}

\section{Baseline Adaptation Protocol}
\label{appendix-baseline-adaptation-protocol}

This section presents the detailed procedure for making a fair comparison between our meta-learning-based method(s) and the baselines under their problem setup difference. Baselines are designed for the \textit{pooled} setting, where each method accesses the entire batch of (meta-)test tasks simultaneously, denoted by $\mathcal{D}_{\text{test}} = \{ D_1', \dots, D_{T'}' \}$. 
In contrast, our \textbf{MetaCaDI} operates in a \textit{sequential} manner, adapting to a single meta-test task without access to other, concurrent test tasks. 

For a fair comparison, we evaluate the test performance by enforcing a strict single-task inference protocol: All methods use only the single (meta-)test dataset for the specific test task being evaluated, without access to the full pool of test tasks.

To fairly compare the performance under such a single-task inference protocol, we take two steps. 

First, for the point-estimation methods (JCI-PC, UT-IGSP, and DCDI-G), we measure the uncertainty in their estimates by employing the standard bootstrapping procedure. This bootstrapping procedure requires repeatedly running each method on multiple bootstrap samples, which is computationally prohibitive. For this reason, we used 5 bootstrap samples for DCDI and 20 samples for the other methods.

Second, we enforce the single-task inference protocol for each method as follows.

\paragraph{UT-IGSP and JCI-PC.}
These methods perform joint inference and hence require access to training and test tasks. To adapt them to the sequential setting, we provided the algorithms with all meta-training datasets plus a single meta-test dataset at a time, averaging the results per bootstrap run. For JCI-PC, due to its prohibitive computational time, we subsampled 10 random training contexts per bootstrap run rather than using the full training set.

\paragraph{DCDI-G.}
Standard DCDI jointly learns the graph and intervention targets via continuous optimization. Running this from scratch for every few-shot test task is computationally infeasible and prone to overfitting. We therefore implemented a two-stage strategy: In Stage 1 (\textbf{Graph Learning}), we train DCDI solely on the (meta-)training data to learn the shared causal graph structure $\cg$. In Stage 2 (\textbf{Episodic Adaptation}), for each isolated (meta-)test task $D_{t'}'$, we freeze the graph structure learned in Stage 1 and optimize only the intervention targets (regime-specific parameters). This approach ensures fairness with MetaCaDI-Analytical (which effectively freezes task-shared parameters during adaptation) and prevents DCDI from overfitting structural parameters to the sparse ($N'_{t'} = 10$) test data.

\paragraph{BaCaDI.}
We utilized the publicly available scripts for BaCaDI in its standard multitask setting but adapted the data input to the episodic protocol. Instead of providing the full pool of test datasets, we applied the method to the set of all training datasets plus a single test dataset at a time. This ensures the method operates without information leakage from concurrent test environments, maintaining consistency with the evaluation pipeline used for the other baselines.

\section{MetaCaDI Implementation Details}

\paragraph{Neural Network Architectures.} To parameterize the likelihood model in \cref{eq:scm_main}, we use two multi-layer perceptrons (MLPs) for each variable $X_i$: one for the observational mechanism $f_i$ and another for the feature extractor in the interventional mechanism $f_i^\text{I}$. We use rectified linear unit (ReLU) activation for all hidden layers and set the number of hidden layers and the number of hidden units in each layer as hyperparameters, which were tuned via random search over the value range presented in \cref{tab:sweep_config} (see \cref{appendix-hyperparameter-search} for details).

To formulate the intervention target identifier $g_{\boldsymbol{\eta}}$ in \cref{eq:logit_predictor}, we use an MLP. 
To obtain its input feature vector $\biF_t$, we use an MLP $h_{\boldsymbol{\varsigma}}$ with a permutation-invariant architecture that yields the feature representation invariant to the order of data instances in a support set $D_t^s$ (\cref{subsubsec:logit_parameters}). To the logit outputs of $g_{\boldsymbol{\eta}}$, we applied layer normalization (\texttt{LayerNorm}) to stabilize training. 

\paragraph{Parameter Optimization.} We train the task-shared parameters $\Phi$ of our MetaCaDI-Analytical by performing differentiable sampling with the \textit{straight-through} estimators \citep{bengio2013estimating}. In the forward pass, we evaluate the objective function value by sampling the causal DAG adjacency matrix $\biA\in\{0,1\}^{d \times d}$ and the intervention targets $\bim_t \in \{0,1\}^{d}$, where $\biA = \biPi^\top\biU\biPi$ is drawn by sampling an upper triangular matrix $\biU \in \{0,1\}^{d \times d}$ from the Gumbel Softmax distribution and a permutation matrix $\biPi \in \{0,1\}^{d\times d}$ from the Gumbel-Top-k distribution. To sample these discrete matrices and vectors, we take $\arg \max$ over the sampled continuous relaxations: e.g., for $\biU$, we compute its $(i,j)$ element $U_{i,j} \in \{0,1\}$ by taking $U_{i,j} = \arg \max [1 - \tilde{U}_{i,j}, \tilde{U}_{i,j}]$, where $\tilde{U}_{i,j}$ is the continuous relaxation of $U_{i,j}$ sampled from the Gumbel-Softmax distribution. In the backward pass, we approximately compute the gradients by using such continuous relaxations as $\tilde{U}_{i,j}$.

By employing such gradients with the Adam optimizer \citep{kingma2014adam}, we trained the task-shared parameters $\Phi$. We used a \texttt{ReduceLROnPlateau} learning rate scheduler and an early stopping mechanism, both monitored on the validation set's ELBO loss, to prevent overfitting.
As with the standard Gumbel-Softmax-based optimization, we annealed the Gumbel temperature for the graph sampler from a higher initial value to a lower final value over a specified percentage of the training epochs to encourage exploration early and exploitation later. 
By contrast, for the intervention target sampler, we used a separate, fixed Gumbel temperature. All hyperparameters, including network architectures, learning rates, regularization weights, and temperature schedules, were tuned using random search, as detailed in \cref{appendix-hyperparameter-search}.

Regarding the task-specific parameters $\Upsilon$, which consists of the final linear layer of the interventional mechanism $f_i^I$, we do not use gradient-based optimization for our MetaCaDI-Analytical. Instead, we analytically computed the optimal values of its weights $\hat{\mathbf{w}}_i^{(t)}$ for each task $t$ using a closed-form solution to a Ridge regression problem (see \cref{subsec:modeling}).

\section{Hyperparameter Search and Settings}
\label{appendix-hyperparameter-search}

For each simulation run, we performed a 100-trial random search over the hyperparameter space detailed in Table \ref{tab:sweep_config}. For this search, the meta-training datasets were partitioned into a training set (80\% of the tasks) and a validation set (the remaining 20\% of tasks). In each trial, a model was trained with a randomly sampled configuration on the 80\% training partition. We used early stopping monitored on the validation set's ELBO to determine the optimal number of training epochs for that trial. After all 100 trials were completed, we selected the single best-performing hyperparameter configuration based on the highest Intv-AUPRC achieved on the validation set. The final metrics reported in the main paper were then computed by retraining a model with this best configuration on the full meta-training data and evaluating it on the held-out test set.

\begin{table*}[!ht]
\centering
\caption{Hyperparameter search space for the random search conducted for each simulation run.}
\label{tab:sweep_config}
\begin{tabular}{l|l|l}
\toprule
\textbf{Parameter} & \textbf{Distribution} & \textbf{Range / Values} \\
\midrule
Learning Rate (lr) & Log-Uniform & [1e-4, 5e-2] \\
Weight Decay & Log-Uniform & [1e-5, 1.0] \\
Gradient Clipping Norm & Categorical & \{0.0, 0.5, 1.0, 2.0\} \\
MLP Hidden Layers & Categorical & \{0, 1, 2, 3\} \\
MLP Hidden Units Ratio & Uniform & [0.25, 1.5] \\
Intervention Predictor Hidden Layers & Categorical & \{1, 2, 3, 4\} \\
Intervention Predictor Hidden Units Ratio & Uniform & [0.5, 2.0] \\
Intervention Sparsity Weight ($\lambda_I$) & Uniform & [0.0, 5.0] \\
Residual Independence Weight ($\lambda_H$) & Uniform & [0.0, 10.0] \\
Reconstruction Loss Weight & Log-Uniform & [0.01, 10.0] \\
Graph Sparsity Prior ($p_G$) & Log-Uniform & [1e-4, 0.1] \\
LR Plateau Patience & Quantized Uniform & [15, 30] \\
Early Stopping Patience & Quantized Uniform & [40, 60] \\
LR Plateau Factor & Uniform & [0.1, 0.5] \\
Initial Gumbel Temperature (Graph) & Uniform & [1.0, 5.0] \\
Final Gumbel Temperature (Graph) & Uniform & [0.1, 0.5] \\
Gumbel Anneal Ratio (Graph) & Uniform & [0.7, 0.9] \\
Gumbel Temperature (Intervention) & Uniform & [0.5, 2.0] \\
\midrule
\textbf{MAML-Specific Parameters} & & \\
Inner Loop Learning Rate & Categorical & \{0.005, 0.01, 0.05\} \\
Inner Loop Steps & Categorical & \{1, 3, 5\} \\
\bottomrule
\end{tabular}
\end{table*}



\section{Additional Experiments}
\label{appendix-additional-experiments}

\subsection{Ablation Study on Intervention Target Predictor}
\label{appendix-ablation}

To validate the effect of the feature and architectural design of the intervention target predictor described in \cref{subsec:modeling}, we performed an ablation study to isolate the contribution of two key design choices: (1) the engineering of the nine-component input feature vector $\biF_t$, and (2) the permutation-invariant Deep Sets architecture.

\subsubsection{Input Feature Ablations}
We assessed the effect of the engineered features by training variants of MetaCaDI with reduced input sets:
\begin{itemize}
    \item \textbf{Raw Data Only (Ablation 1A):} The model receives only the support set data $D_t^s$, excluding all predicted values and residuals.
    \item \textbf{No Residuals (Ablation 1B):} The model receives the support set data and statistical summaries but explicitly excludes the predicted values ($\hat{\mathbf{X}}^s_\text{obs}, \hat{\mathbf{X}}^s_\text{int}$) and their residuals. 
    \item \textbf{No Statistical Summaries (Ablation 1C):} The model uses all per-sample features (including residuals) but excludes the broadcasted mean and standard deviation.
\end{itemize}

\subsubsection{Architecture Ablations}
We evaluated the utility of the Deep Sets architecture against simpler aggregation methods:
\begin{itemize}
    \item \textbf{Naive Mean Pooling (Ablation 2A):} The Deep Sets architecture is replaced by a simple column-wise mean of the feature matrix $\mathbf{C}_t$.
    \item \textbf{Single-Branch Pooling (Ablation 2B):} We tested using only the ``Mean of Embeddings'' branch ($\mathbf{z}_{\text{ME}}$) or only the ``Embedding of Means'' branch ($\mathbf{z}_{\text{EM}}$), rather than their concatenation.
    \item \textbf{Flattened MLP (Ablation 2C):} We removed the permutation invariance constraint by flattening the feature matrix $\mathbf{C}_t$ into a single vector and feeding it into a standard MLP.
\end{itemize}

\subsubsection{Results and Analysis}
We summarize the results on both SERGIO and Synthetic datasets in \Cref{tab:ablation_results}.

\textbf{Necessity of Residuals.} The ``base'' MetaCaDI-Analytical model significantly outperforms Ablation 1B (No Residuals). On the Synthetic dataset, removing residuals caused the Intv-AUPRC to drop from $0.351$ to $0.233$. This confirms that the model relies heavily on the \textit{residuals}---the discrepancy between the observed data and the SCM predictions---to identify intervention targets, rather than simply inferring them from the raw data distribution.

\textbf{Impact on Graph Recovery.} Crucially, the degradation in the intervention identifier leads to a collapse in graph recovery performance. For example, in the Raw Data Only ablation (1A), the E-SID on Synthetic data worsened from $47.88$ to $109.47$. This highlights the tight coupling in our joint inference framework: accurate intervention target identification is a prerequisite for reliable causal discovery.

\textbf{Benefits of Deep Sets.} The architectural ablations demonstrate that the Deep Sets formulation is superior to Naive Mean Pooling (2A) and Flattened MLP (2C). The drop in performance for the Flattened MLP suggests that enforcing permutation invariance provides a strong inductive bias that prevents overfitting to the specific ordering of the few-shot support set.

\begin{table*}[!ht]
\centering
\caption{Ablation study results on SERGIO and Synthetic datasets. \textbf{Base} refers to the proposed MetaCaDI-Analytical method. Results are reported as mean $\pm$ standard deviation over 20 simulations. The best results are bolded. $\downarrow$ and $\uparrow$ indicate ``lower is better'' and ``higher is better'', respectively.}
\label{tab:ablation_results}
\resizebox{\textwidth}{!}{%
\begin{tabular}{l|cccc|cccc}
\toprule
& \multicolumn{4}{c|}{\textbf{SERGIO Dataset}} & \multicolumn{4}{c}{\textbf{Synthetic Dataset}} \\
\textbf{Method} & \textbf{Intv-AUROC} $\uparrow$ & \textbf{Intv-AUPRC} $\uparrow$ & \textbf{E-SHD} $\downarrow$ & \textbf{E-SID} $\downarrow$ & \textbf{Intv-AUROC} $\uparrow$ & \textbf{Intv-AUPRC} $\uparrow$ & \textbf{E-SHD} $\downarrow$ & \textbf{E-SID} $\downarrow$ \\
\midrule
\textbf{MetaCaDI-Analytical} & \textbf{0.740} $\pm$ \textbf{0.038} & \textbf{0.094} $\pm$ \textbf{0.007} & \textbf{39.39} $\pm$ \textbf{1.51} & \textbf{70.87} $\pm$ \textbf{10.32} & \textbf{0.604} $\pm$ \textbf{0.045} & \textbf{0.351} $\pm$ \textbf{0.081} & \textbf{26.63} $\pm$ \textbf{7.47} & \textbf{47.88} $\pm$ \textbf{20.44} \\
\midrule
\textit{Feature Ablations} & & & & & & & & \\
1A: Raw Data Only & 0.668 $\pm$ 0.082 & 0.078 $\pm$ 0.015 & 57.16 $\pm$ 23.53 & 100.49 $\pm$ 38.16 & 0.545 $\pm$ 0.040 & 0.242 $\pm$ 0.041 & 64.08 $\pm$ 35.73 & 109.47 $\pm$ 59.18 \\
1B: No Residuals & 0.688 $\pm$ 0.063 & 0.083 $\pm$ 0.013 & 57.32 $\pm$ 23.49 & 100.70 $\pm$ 37.85 & 0.537 $\pm$ 0.026 & 0.233 $\pm$ 0.019 & 63.75 $\pm$ 35.62 & 108.98 $\pm$ 59.16 \\
1C: No Stat. Summaries & 0.676 $\pm$ 0.087 & 0.085 $\pm$ 0.017 & 56.89 $\pm$ 23.34 & 100.06 $\pm$ 37.93 & 0.546 $\pm$ 0.040 & 0.244 $\pm$ 0.037 & 63.91 $\pm$ 35.58 & 109.21 $\pm$ 58.93 \\
\midrule
\textit{Architecture Ablations} & & & & & & & & \\
2A: Naive Mean Pool & 0.675 $\pm$ 0.072 & 0.086 $\pm$ 0.016 & 57.15 $\pm$ 23.27 & 100.57 $\pm$ 37.78 & 0.581 $\pm$ 0.038 & 0.299 $\pm$ 0.061 & 64.17 $\pm$ 35.58 & 109.95 $\pm$ 59.08 \\
2Bi: Only $\mathbf{z}_{\text{ME}}$ & 0.676 $\pm$ 0.068 & 0.081 $\pm$ 0.012 & 57.44 $\pm$ 23.53 & 101.04 $\pm$ 37.80 & 0.540 $\pm$ 0.035 & 0.235 $\pm$ 0.032 & 64.24 $\pm$ 35.73 & 109.71 $\pm$ 59.25 \\
2Bii: Only $\mathbf{z}_{\text{EM}}$ & 0.680 $\pm$ 0.064 & 0.082 $\pm$ 0.013 & 57.43 $\pm$ 23.57 & 101.32 $\pm$ 38.08 & 0.545 $\pm$ 0.037 & 0.240 $\pm$ 0.034 & 63.97 $\pm$ 35.60 & 109.33 $\pm$ 58.90 \\
2C: Flattened MLP & 0.666 $\pm$ 0.082 & 0.086 $\pm$ 0.023 & 56.94 $\pm$ 23.34 & 99.84 $\pm$ 37.76 & 0.567 $\pm$ 0.037 & 0.287 $\pm$ 0.052 & 63.97 $\pm$ 35.58 & 109.59 $\pm$ 59.18 \\
\bottomrule
\end{tabular}
}
\end{table*}

\subsection{Computational Runtime Comparison}
\label{appendix-runtime-comparison}

\begin{figure*}[htpb]
\centering
\includegraphics[width=\textwidth]{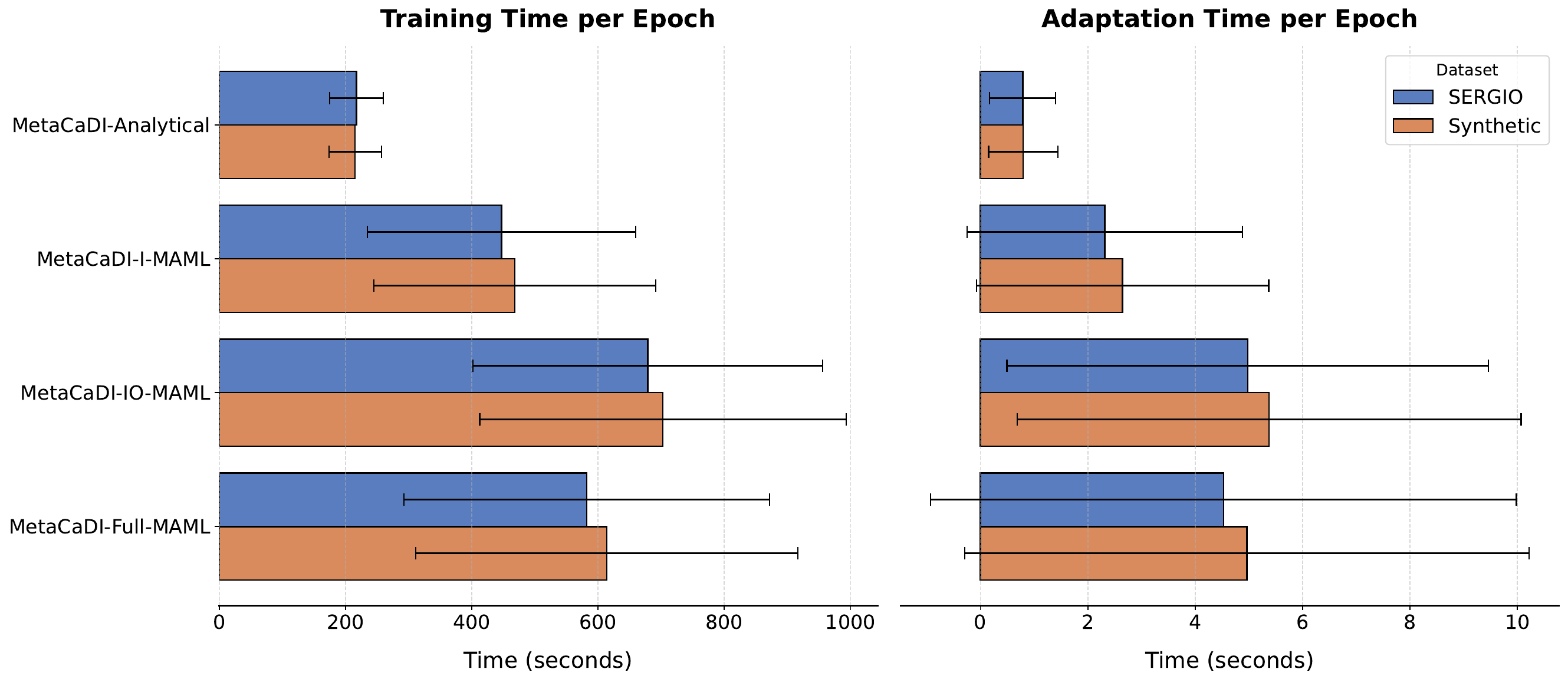}
\caption{Per-epoch runtime analysis for MetaCaDI variants. The plots compare the average training time (left) and adaptation time (right) per epoch on the SERGIO and Synthetic datasets. Error bars represent the standard deviation over 20 simulations.}
\label{fig:per_epoch_runtime}
\end{figure*}

\begin{figure}[!ht]
\centering
\includegraphics[width=0.8\linewidth]{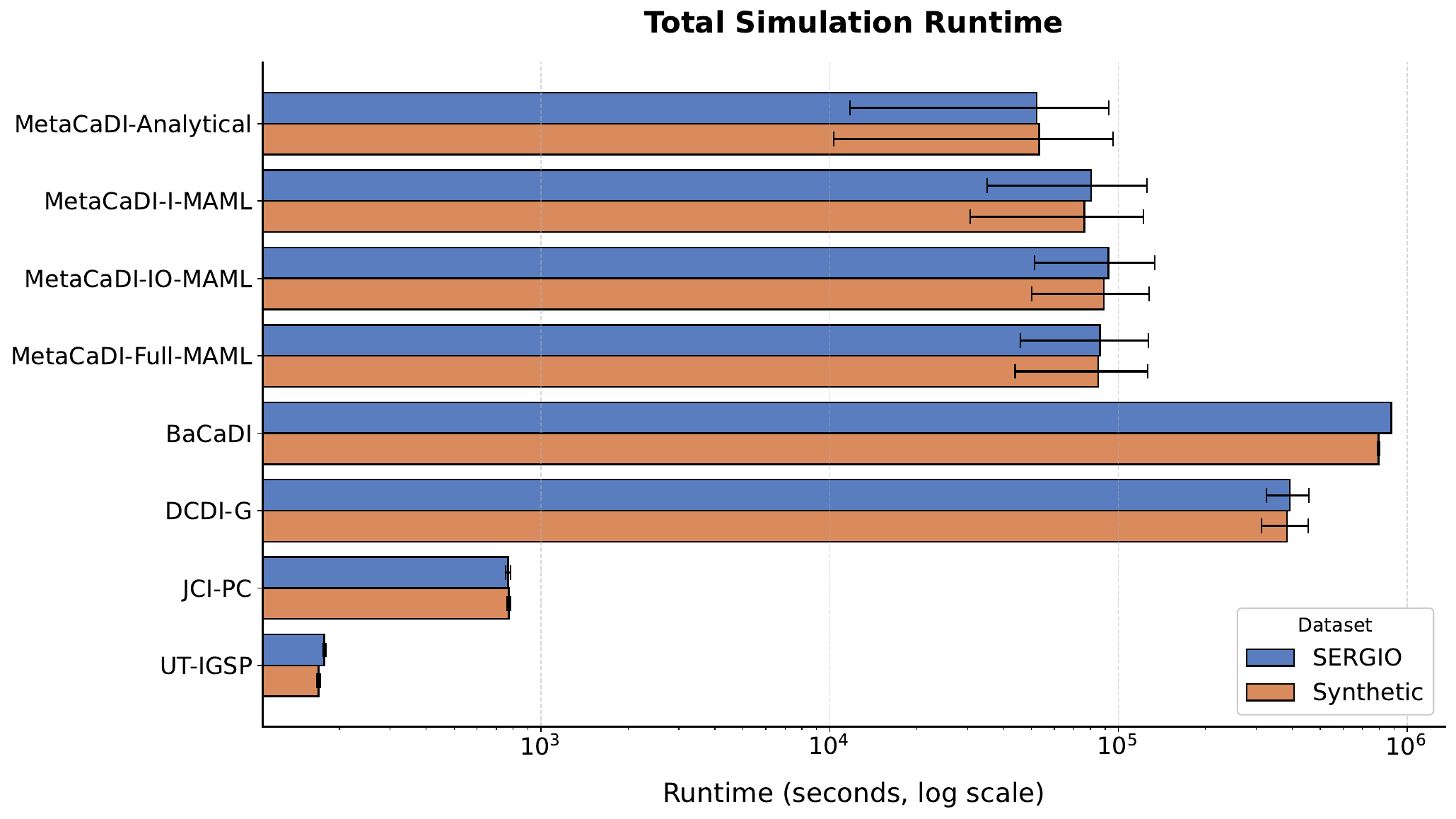}
\caption{Average total runtime per dataset. The plot compares the time required to learn both the shared causal graph and the intervention targets for each of the 20 test tasks. Note that the x-axis uses a logarithmic scale to accommodate the wide range of values. Error bars represent the standard deviation.}
\label{fig:total_runtime}
\end{figure}

We present runtime from two perspectives: \Cref{fig:per_epoch_runtime} compares our proposed method against its MAML-based variants on a per-epoch basis for a precise analysis of the efficiency gains from our analytical adaptation strategy. \Cref{fig:total_runtime} shows the total runtime per dataset in order to make it comparable with the baselines.

The per-epoch results in \Cref{fig:per_epoch_runtime} clearly demonstrate the efficiency of the analytical approach. On the SERGIO dataset, MetaCaDI-Analytical is approximately \textbf{2.7 times faster} per training epoch than MetaCaDI-Full-MAML ($217.59$ vs. $581.95$ [sec.]). The benefit is even more pronounced in the task-adaptation step, where MetaCaDI-Analytical is approximately \textbf{5.7 times faster} than MetaCaDI-Full-MAML ($0.79$ vs. $4.53$ [sec.]).

In terms of total runtime per dataset, MetaCaDI-Analytical demonstrates superior scalability compared to other continuous optimization and Bayesian frameworks. BaCaDI is the most computationally intensive, requiring an average of $\sim 245$ hours to complete. DCDI-G also incurs a high cost, averaging $\sim 109$ hours. In contrast, MetaCaDI-Analytical completes the full pipeline for a dataset in $\sim 14.5$ hours, offering an order-of-magnitude speedup over comparable continuous baselines.

\subsection{Sensitivity to Number of Variables $d$}
\label{appendix-sensitivity-d}

To further characterize the behavior of MetaCaDI, we conducted a sensitivity analysis on both the SERGIO and Synthetic datasets by varying the number of variables $d \in \{10, 20, 30\}$.

\paragraph{Experimental Setup.}
To ensure a rigorous comparison, we adhered to the data generation and evaluation protocols described in \Cref{sec:experiments} and \Cref{appendix-data-gen}.
For all settings ($d \in \{10, 20, 30\}$), we maintained the same sample size ($N_t = 110$) and meta-training/test splits.
Crucially, to maintain a consistent level of difficulty relative to the system size, the number of ground-truth intervention targets in each environment was scaled proportionally to the number of nodes.
Specifically, we targeted 20\% of the variables in each experimental setting (i.e., 2, 4, and 6 targets for $d=10, 20, \text{and } 30$, respectively).

\begin{figure*}[!h]
\centering
\includegraphics[width=\textwidth]{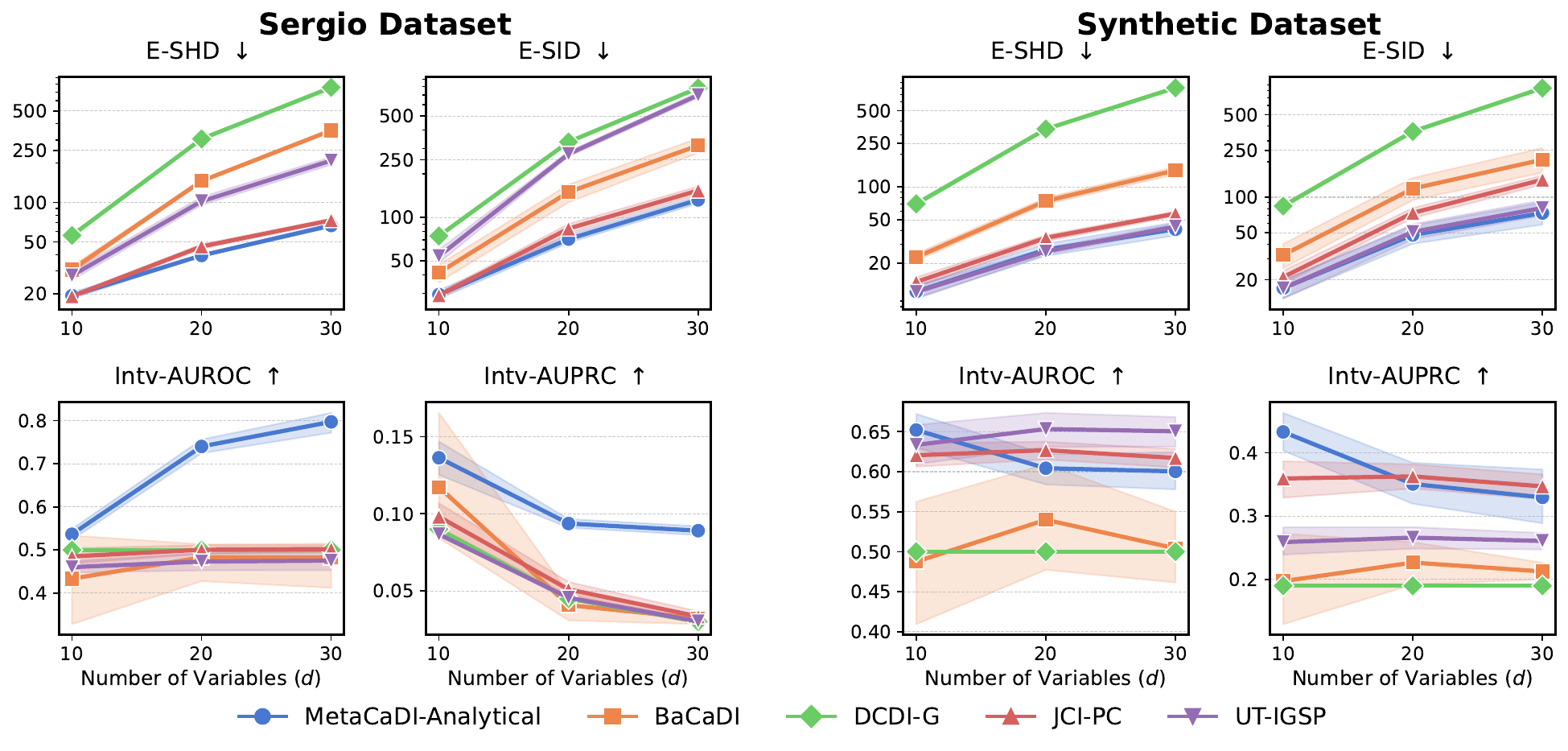}
\caption{Performance sensitivity to graph size ($d$) on the SERGIO and synthetic datasets.
Results are averaged over 20 independent simulations.
MetaCaDI-Analytical (blue circles) is compared against baselines across $d \in \{10, 20, 30\}$.
$\downarrow$ and $\uparrow$ indicate ``lower is better'' and ``higher is better'', respectively.}
\label{fig:sensitivity_performance}
\end{figure*}

\paragraph{Performance Trends.}
The quantitative results are visualized in \Cref{fig:sensitivity_performance}. 
On the SERGIO dataset, we observe a divergence between the two intervention identification metrics.
The Intv-AUROC performance of MetaCaDI-Analytical improves as the system size increases, rising from $0.536$ at $d=10$ to $0.797$ at $d=30$. While Intv-AUPRC does not show the same increasing trend, our MetaCaDI-Analytical still outperforms all baselines.


Regarding graph structure recovery, we observe that error metrics (E-SHD and E-SID) naturally increase with the super-linear growth of the graph search space.
For instance, at $d=30$, MetaCaDI achieves an E-SHD of $66.7$, whereas BaCaDI and DCDI-G degrade significantly to $355.3$ and $756.0$, respectively. We note that MetaCaDI-Analytical performs competitively in all these cases.

\begin{figure}[!ht]
\centering
\includegraphics[width=0.9\textwidth]{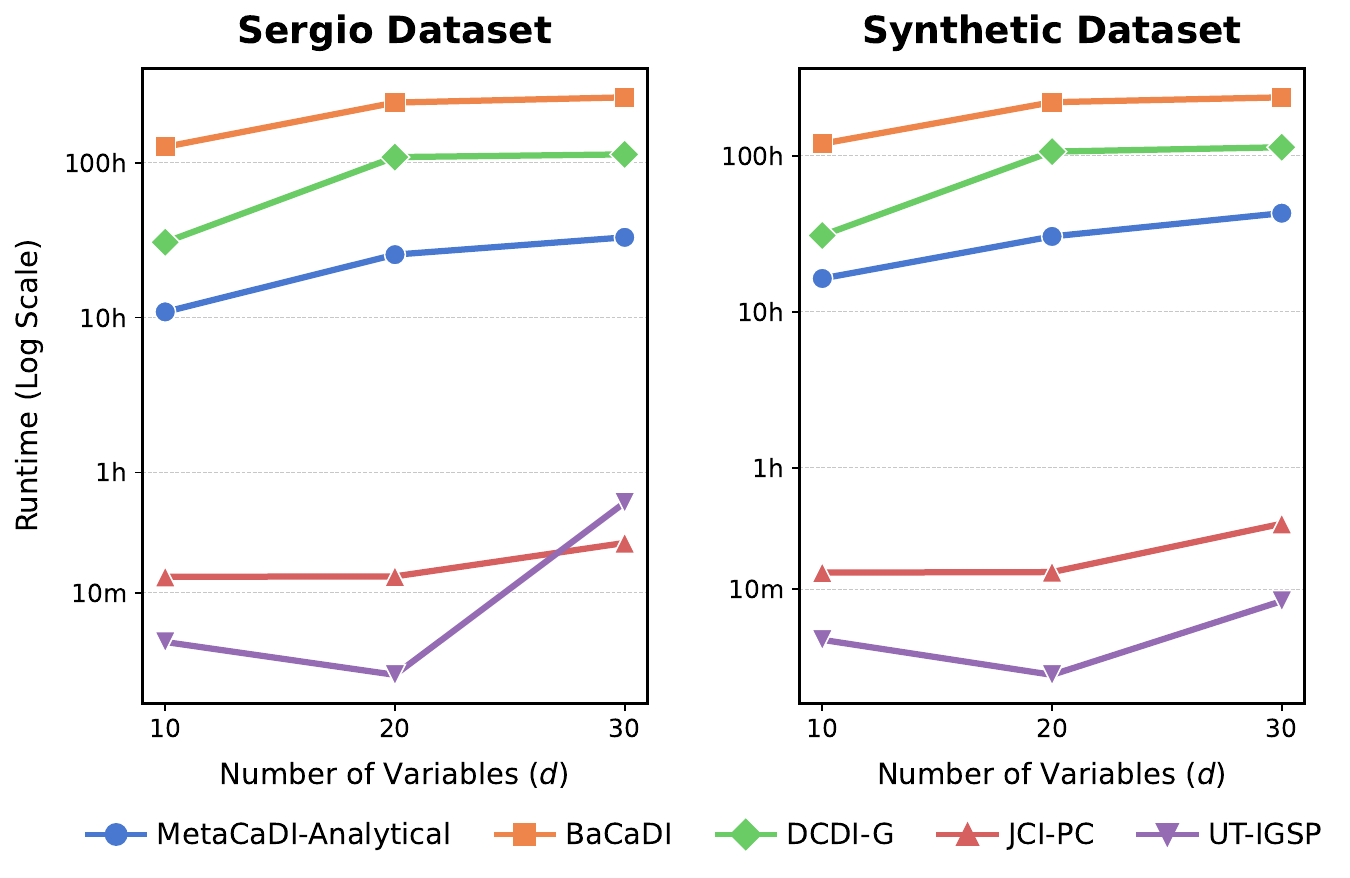}
\caption{\textbf{Runtime scalability ($d$).} 
The total runtime (training + evaluation) is plotted on a logarithmic scale. 
MetaCaDI-Analytical offers significant speedups over BaCaDI and DCDI-G, maintaining manageable runtimes even as the system size scales.}
\label{fig:sensitivity_runtime}
\end{figure}

\paragraph{Runtime Scalability.}
A key advantage of our proposed analytical adaptation is its computational efficiency. 
\Cref{fig:sensitivity_runtime} illustrates the total runtime (in hours) required for a full simulation run.
MetaCaDI-Analytical exhibits superior scalability compared to the other continuous optimization frameworks.
For the largest setting ($d=30$) on the SERGIO dataset, MetaCaDI completes the full pipeline in approximately 33.0 hours. 
In stark contrast, DCDI-G requires 113.3 hours ($3.4\times$ slower), and BaCaDI requires 264.6 hours ($8\times$ slower).
While the constraint-based (JCI-PC) and score-based (UT-IGSP) methods are faster due to their combinatorial nature on small graphs, they often fail to achieve comparable accuracy on the complex, non-linear SERGIO benchmarks.
These results confirm that MetaCaDI provides an optimal balance, offering the high expressivity of neural networks without the prohibitive computational costs typically associated with bi-level optimization in larger causal graphs.

\subsection{Few-Shot Robustness}
\label{appendix-sensitivity-k}

\begin{figure*}[!ht]
\centering
\includegraphics[width=\textwidth]{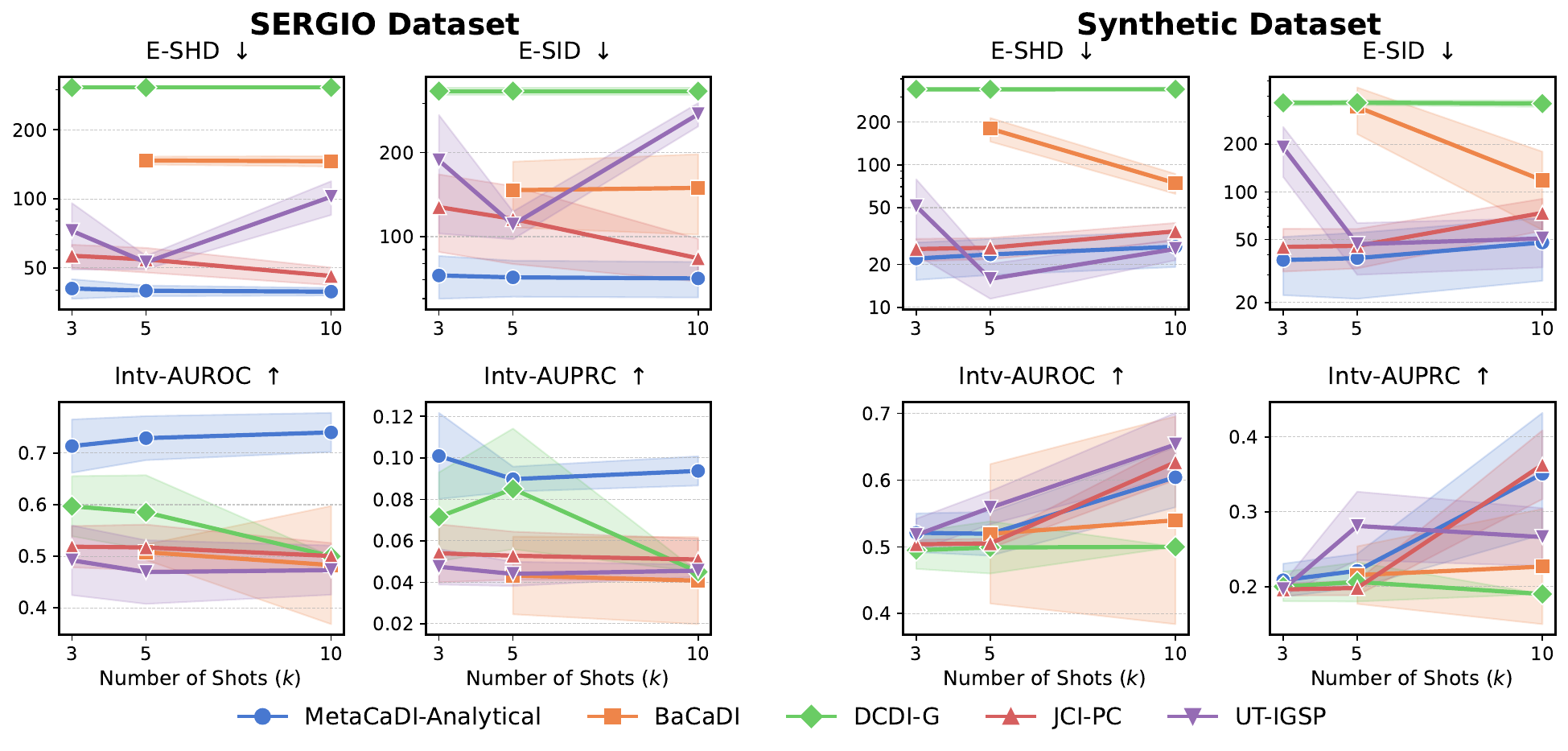}
\caption{\textbf{Performance sensitivity to support set size ($k$).} 
The plots display the mean and standard deviation of performance metrics over 20 independent simulations for varying shot counts $k \in \{3, 5, 10\}$ on the SERGIO (left) and Synthetic (right) datasets.}
\label{fig:sensitivity_k}
\end{figure*}

To assess the robustness of MetaCaDI in the extreme low-data regime, we evaluated performance by varying the support set size $k \in \{3, 5, 10\}$ used for identifying intervention targets in the meta-test tasks. The results on both the SERGIO and Synthetic datasets are visualized in \Cref{fig:sensitivity_k}.

\paragraph{Robustness in Complex Environments (SERGIO).}
On the realistic SERGIO dataset, MetaCaDI demonstrates exceptional stability. Crucially, it consistently achieves the highest Intv-AUROC and Intv-AUPRC and the lowest structural errors (E-SHD, E-SID) across all evaluated sample sizes ($k$).
\begin{itemize}
    \item \textbf{Intervention Identification:} At $k=3$, MetaCaDI achieves an Intv-AUROC of $0.714 \pm 0.052$ and an Intv-AUPRC of $0.101 \pm 0.021$. In contrast, most baseline methods fail to extract meaningful signals in this ultra-low data regime. While DCDI-G performs slightly better than random chance (Intv-AUROC $\approx 0.597$), it still lags significantly behind our method. \textbf{Notably, BaCaDI fails to adapt to this extreme sparsity,} and JCI-PC ($\approx 0.519$) similarly yields results close to random guess. This gap confirms that our meta-learning framework successfully leverages shared knowledge to regularize inference when task-specific data is insufficient for standalone discovery.
    \item \textbf{Structural Stability:} The structural error metrics for MetaCaDI remain remarkably consistent across sample sizes (E-SHD $\approx 40$, E-SID $\approx 71$ for all $k$). Conversely, baselines exhibit high sensitivity or instability. JCI-PC degrades from an E-SHD of $46.1$ ($k=10$) to $56.5$ ($k=3$), while UT-IGSP shows erratic behavior, with E-SHD fluctuating from $102.2$ ($k=10$) to $52.9$ ($k=5$) and back to $72.5$ ($k=3$).
\end{itemize}

\paragraph{Threshold Effects in Synthetic Data.}
In the Synthetic setting, we observe a distinct performance threshold for intervention identification.
\begin{itemize}
    \item \textbf{Sample Efficiency:} At $k=3$ and $k=5$, both MetaCaDI and the baselines struggle to identify intervention targets with high precision, with Intv-AUPRC scores hovering around $0.20-0.22$. However, at $k=10$, MetaCaDI exhibits a sharp performance improvement, reaching an Intv-AUPRC of $0.351 \pm 0.081$. This behavior mirrors that of the strongest baseline, JCI-PC, which also sees a similar increase to $0.363 \pm 0.046$ at $k=10$.
    \item \textbf{Interpretation:} This suggests that in these synthetic environments, a minimum sample size (around $k=10$) is likely required for the statistical signatures of interventions to become distinguishable from observational noise. Once this threshold is met, MetaCaDI remains competitive with the best-performing baselines.
\end{itemize}

\paragraph{Summary.}
These results highlight a key advantage of MetaCaDI: while it performs competitively to standard methods in idealized settings with sufficient data ($k=10$, Synthetic), it provides a critical safety net in complex, data-scarce scenarios ($k=3$, SERGIO), where it is the only method capable of reliable inference.

\subsection{Robustness to Environmental Distribution Shifts}
\label{appendix-distribution-shifts}

To systematically evaluate robustness to distribution shifts, we modify the Structural Causal Model (SCM) generation process for the unseen meta-test tasks. Let $X_j = f_j(\text{Pa}_j) + \epsilon_j$ be the base SCM for a given environment. We apply the following perturbations during the meta-test phase:
\begin{enumerate}[leftmargin=*]
    \item \textbf{Task-Dependent Noise Variances ($\sigma$):} The exogenous noise is scaled such that the generated observational data follows $X_j = f_j(\text{Pa}_j) + (\sigma \cdot \epsilon_j)$, where $\sigma \sim \text{Uniform}(0.5, 3.0)$. This tests the algorithm's ability to recover structures when the signal-to-noise ratio drastically changes.
    \item \textbf{Causal Mechanism Drift ($\delta$):} The functional mapping from parents to children is altered such that $X_j = (\delta \cdot f_j(\text{Pa}_j)) + \epsilon_j$, where $\delta \sim \text{Uniform}(0.8, 1.2)$. This tests robustness against deterministic changes in the underlying causal dynamics.
\end{enumerate}

\Cref{tab:distribution_shifts} summarizes the performance of MetaCaDI against the baselines under these distribution shifts. MetaCaDI-Analytical maintains highly robust structural recovery (achieving the lowest E-SHD) despite the environmental drift. The closed-form analytical solver used for task adaptation effectively acts as a regularized linear layer over the deeply learned shared features, preventing the model from overfitting to the shifted noise or drifting mechanisms. 

In contrast, standard continuous multi-task approaches (BaCaDI and DCDI-G) exhibit catastrophic structural degradation under both shifts, suffering from excessively high E-SHD scores as they fail to disentangle the shifted environments from the underlying causal structure. While the score-based UT-IGSP demonstrated partial robustness to noise variance shifts---performing competitively in intervention identification---it degraded significantly under the more complex causal mechanism drift, a scenario where MetaCaDI continued to dominate across all metrics.

\begin{table}[!ht]
\centering
\caption{Performance under environmental distribution shifts on the Synthetic dataset ($d=20$, $k=10$).}
\label{tab:distribution_shifts}
\begin{tabular}{lccc}
\toprule
\textbf{Method} & \textbf{Intv-AUPRC} $\uparrow$ & \textbf{Intv-AUROC} $\uparrow$ & \textbf{E-SHD} $\downarrow$ \\
\midrule
\multicolumn{4}{c}{\textit{Shift 1: Task-Dependent Noise Variances ($\sigma \sim \text{Uniform}(0.5, 3.0)$)}} \\
\midrule
MetaCaDI-Analytical & $\mathbf{0.293 \pm 0.080}$ & $0.592 \pm 0.065$ & $\mathbf{8.845 \pm 3.668}$ \\
BaCaDI & $0.187 \pm 0.055$ & $0.462 \pm 0.110$ & $138.559 \pm 8.013$ \\
DCDI-G & $0.190 \pm 0.000$ & $0.500 \pm 0.000$ & $367.700 \pm 4.169$ \\
JCI-PC & $0.215 \pm 0.017$ & $0.552 \pm 0.027$ & $38.490 \pm 5.921$ \\
UT-IGSP & $0.276 \pm 0.033$ & $\mathbf{0.647 \pm 0.037}$ & $11.980 \pm 2.303$ \\
\midrule
\multicolumn{4}{c}{\textit{Shift 2: Causal Mechanism Drift ($\delta \sim \text{Uniform}(0.8, 1.2)$)}} \\
\midrule
MetaCaDI-Analytical & $\mathbf{0.253 \pm 0.044}$ & $\mathbf{0.578 \pm 0.041}$ & $\mathbf{7.521 \pm 2.563}$ \\
BaCaDI & $0.211 \pm 0.035$ & $0.506 \pm 0.110$ & $136.737 \pm 7.869$ \\
DCDI-G & $0.190 \pm 0.000$ & $0.500 \pm 0.000$ & $368.300 \pm 4.692$ \\
JCI-PC & $0.197 \pm 0.009$ & $0.507 \pm 0.024$ & $47.250 \pm 9.951$ \\
UT-IGSP & $0.201 \pm 0.020$ & $0.524 \pm 0.048$ & $12.680 \pm 3.097$ \\
\bottomrule
\end{tabular}%
\end{table}

\subsection{Performance on Linear Gaussian Data}
\label{appendix-linear-gaussian}

To evaluate the robustness of MetaCaDI against model misspecification, we assess its performance on data generated from linear Gaussian ANMs. Unlike non-linear ANMs, linear Gaussian models lack functional identifiability from observational data alone (\cref{rem:identifiability}). Consequently, the model cannot exploit functional signatures to determine causal directionality and must rely strictly on interventional signals (i.e., distributional shifts across regimes) to recover the causal structure. Because MetaCaDI parameterizes mechanisms using non-linear neural networks, this setting serves as a challenging test case where the model's inductive biases diverge from the underlying data-generating process.

Under linear Gaussian assumptions, the causal graph is identifiable only up to I-MEC. Therefore, directly computing the SHD between the inferred and true DAGs is inappropriate. Instead, we evaluate structural recovery by converting each DAG into an Interventional Completed Partially Directed Acyclic Graph (I-CPDAG) and computing the expected SHD between these I-CPDAGs to accurately reflect the distance between I-MECs.

\Cref{tab:linear_results} summarizes the results. Despite lacking functional identifiability guarantees in this setting, MetaCaDI demonstrates robust causal structure recovery, achieving the lowest E-SHD ($21.307$) and significantly outperforming baseline continuous optimization methods such as BaCaDI ($96.100$) and DCDI-G ($170.550$). This confirms that MetaCaDI effectively isolates and exploits intervention signals even in non-ANM settings. 

For intervention target identification, the score-based UT-IGSP achieves the highest performance ($\text{Intv-AUROC} = 0.628$, $\text{Intv-AUPRC} = 0.292$). Nevertheless, MetaCaDI remains highly competitive ($\text{Intv-AUROC} = 0.527$, $\text{Intv-AUPRC} = 0.233$), ranking second and outperforming the remaining baselines. 

\begin{table}[!ht]
\centering
\caption{Performance on Linear Gaussian Data ($d=20$, averaged over 20 simulations). Structural recovery is measured via the expected SHD between I-CPDAGs to account for I-MEC identifiability limits.}
\label{tab:linear_results}
\begin{tabular}{lccc}
\toprule
\textbf{Method} & \textbf{Intv-AUPRC} $\uparrow$ & \textbf{Intv-AUROC} $\uparrow$ & \textbf{E-SHD} $\downarrow$ \\
\midrule
MetaCaDI-Analytical & $0.233 \pm 0.046$ & $0.527 \pm 0.040$ & $\mathbf{21.307 \pm 6.839}$ \\
BaCaDI & $0.206 \pm 0.067$ & $0.510 \pm 0.126$ & $96.100 \pm 14.828$ \\
DCDI-G & $0.190 \pm 0.000$ & $0.500 \pm 0.000$ & $170.550 \pm 5.463$ \\
JCI-PC & $0.196 \pm 0.010$ & $0.504 \pm 0.008$ & $25.610 \pm 4.609$ \\
UT-IGSP & $\mathbf{0.292 \pm 0.053}$ & $\mathbf{0.628 \pm 0.050}$ & $24.870 \pm 12.453$ \\
\bottomrule
\end{tabular}%
\end{table}

\end{document}